%% file: arxiv.tex
\documentclass[10pt,twocolumn,letterpaper]{article}

\usepackage[pagenumbers]{cvpr} 

\input{preamble}
\input{my_preamble}
\input{alphabet}

\definecolor{cvprblue}{rgb}{0.21,0.49,0.74}
\usepackage[pagebackref,breaklinks,colorlinks,allcolors=cvprblue]{hyperref}

\def\paperID{} 
\def\confName{CVPR}
\def\confYear{2026}

\title{PyramidalWan:\\ On Making Pretrained Video Model Pyramidal for Efficient Inference}

\author{%
Denis Korzhenkov\thanks{Equal contribution}%
\quad Adil Karjauv\footnotemark[1] %
\quad Animesh Karnewar %
\quad Mohsen Ghafoorian %
\quad Amirhossein Habibian \\
{Qualcomm AI Research\thanks{Qualcomm AI Research is an initiative of Qualcomm Technologies, Inc. Snapdragon and Qualcomm branded products are products of Qualcomm Technologies, Inc. and/or its subsidiaries.} } \\
{\tt\small \{dkorzhen, akarjauv, karnewar, mghafoor, ahabibia\}@qti.qualcomm.com} 
}

\begin{document}
\maketitle
\input{sec/0_abstract}    
\input{sec/1_intro}

\input{sec/2_related}

\input{sec/3_method}
\input{sec/4_experiments}

\input{sec/5_conclusion}
{
    \small
    \bibliographystyle{ieeenat_fullname}
    \bibliography{main}
}

\input{supplementary/X_suppl}
\input{supplementary/wavelets}
\input{supplementary/experiments}
\input{supplementary/details}
\end{document}

%% file: preamble.tex
\renewcommand{\paragraph}[1]{\vspace{.5em}\noindent\textbf{#1.}}

\usepackage{xspace}

\usepackage{soul}
\setuldepth{foobar}

%% file: my_preamble.tex
\usepackage[utf8]{inputenc} 
\usepackage[T1]{fontenc}    
\usepackage{url}            
\usepackage{nicefrac}       
\usepackage{xcolor}         

\usepackage{booktabs}       
\usepackage{multirow}
\usepackage{makecell}

\usepackage{algorithm}
\usepackage{algorithmic}

\usepackage{enumitem}
\usepackage[inline]{enumitem}  
\usepackage{subcaption}
\usepackage{wrapfig}
\usepackage[percent]{overpic} 

\usepackage[group-separator={,},group-minimum-digits={3}]{siunitx}

\usepackage{pifont}  

\usepackage{amsmath}
\usepackage{amsfonts}       
\usepackage{amssymb}
\usepackage{mathtools}
\usepackage{amsthm}
\theoremstyle{plain}

\theoremstyle{definition}

\theoremstyle{remark}

\newcommand\lft{\mathopen{}\left}
\newcommand\rgt{\aftergroup\mathclose\aftergroup{\aftergroup}\right}

\newcommand{\oneincircle}{\ding{172}\xspace}
\newcommand{\twoincircle}{\ding{173}\xspace}

\newcommand{\yesmark}{\ding{51}}%
\newcommand{\nomark}{\ding{55}}%

%% file: alphabet.tex
\usepackage{ifthen}

\newcommand{\snrlevel}[1][]{%
    \ifthenelse{ \equal{#1}{} }
    {\varsigma}
    {\varsigma^{\left(#1\right)}}
}

\newcommand{\globlevel}[1][]{%
    \ifthenelse{ \equal{#1}{} }
    {\sigma}
    {\sigma^{\left(#1\right)}}
}

\newcommand{\loclevel}[1][]{\ensuremath{%
    \ifthenelse{ \equal{#1}{} }
    {\rho}
    {\rho^{\left(#1\right)}}
}}

\newcommand{\hi}{\ensuremath{\textrm{hi}}}
\newcommand{\lo}{\ensuremath{\textrm{lo}}}
\newcommand{\x}{\ensuremath{\textrm{x}}}
\newcommand{\y}{\ensuremath{\textrm{y}}}
\newcommand{\downsample}{\ensuremath{\mathfrak{R}^{\downarrow}}}
\newcommand{\upsample}{\ensuremath{\mathfrak{R}^{\uparrow}}}
\newcommand{\upsamplegauss}{\ensuremath{\mathfrak{R}^{\uparrow}_{\mathcal{N}}}}
\newcommand{\upi}{\ensuremath{\left(i\right)}}
\newcommand{\upiplus}{\ensuremath{\left(i+1\right)}}

%% file: sec/0_abstract.tex
\begin{abstract}
Recently proposed pyramidal models decompose the conventional forward and backward diffusion processes into multiple stages operating at varying resolutions. 
These models handle inputs with higher noise levels at lower resolutions, while less noisy inputs are processed at higher resolutions. 
This hierarchical approach significantly reduces the computational cost of inference in multi-step denoising models. 
However, existing open-source pyramidal video models have been trained from scratch and tend to underperform compared to state-of-the-art systems in terms of visual plausibility.
In this work, we present a pipeline that converts a pretrained diffusion model into a pyramidal one through low-cost finetuning, achieving this transformation without degradation in quality of output videos.
Furthermore, we investigate and compare various strategies for step distillation within pyramidal models, aiming to further enhance the inference efficiency.
Our results are available at \url{https://qualcomm-ai-research.github.io/PyramidalWan}
\end{abstract}
\vspace{-1em}

%% file: sec/1_intro.tex
\section{Introduction}
\label{sec:intro}

\begin{table}[t]
\centering
\caption{\textbf{Computational costs.} Schedule is the number of steps per each of three stages, from the lowest spatiotemporal resolution to the highest. For diffusion models, the cost is doubled due to the usage of classifier-free guidance.}
\label{tab:flops}
\resizebox{0.75\linewidth}{!}{%
\input{tables/flops}
}
\vspace{0.7em}
\caption{\textbf{Latency of a single denoiser forward pass.} 
The savings obtained due to the reduced number of tokens at stages 1 and 2 lead to 43\% speedup for 2-2-1 schedule in comparison with 0-0-2 while being only 13\% slower than 0-0-1.
Video DiT was compiled for each stage-wise resolution separately for this measurement.
}
\label{tab:latency}
\resizebox{0.75\linewidth}{!}{%
\input{tables/latency}
}
\vspace{-1.25em}
\end{table}

Recent video diffusion models have achieved remarkable generative quality~\cite{luminavideo,kong2025hunyuanvideosystematicframeworklarge,yang2025cogvideox,HaCohen2024LTXVideo}. 
However, these impressive capabilities come at a cost: multi-step inference remains computationally expensive. 
The principal strategies for reducing inference overhead are step distillation and architectural optimization~\cite{pmlr-v267-lin25m,chen2025sanavideoefficientvideogeneration,zhang2025fast,zhang2025faster,ghafoorian2025attentionsurgeryefficientrecipe}.

Beyond these approaches, several recent works have proposed training diffusion models that process inputs with different noise levels at different resolutions~\cite{chen2025pixelflow,teng2024relay,campbell2023transdimensional,minimax_team_minimax_2025,zhang2025training}. 
This method is motivated by the observation known as \emph{spectral autoregression}: in the spectral decomposition of natural signals, higher-frequency components tend to have lower magnitudes and thus are eliminated earlier during the forward diffusion process~\cite{dieleman2024spectral,falck2025fourierspaceperspectivediffusion,rissanen2023generative}. 
This insight can be exploited to make generation more efficient.
In the beginning, the generator starts with pure low-resolution Gaussian  noise,  synthesizes (still noisy) coarse structure at the same resolution, and then progressively increases the resolution simultaneously with further denoising.
\citet{jin_pyramidal_2025} proposed the formalization of this approach in a form of  \emph{PyramidalFlow} framework.
However, they only demonstrated training of such a model from scratch under limited computational resources.
In our work, we show that pretrained state-of-the-art diffusion models can be made pyramidal via low-cost finetuning  without loss in visual quality.

In detail, we begin with the pretrained Wan2.1-1.3B model~\cite{wan2025wanopenadvancedlargescale} and decompose its forward and backward diffusion processes into three spatiotemporal stages, operating at resolutions of $81 \times 448 \times 832$, $41 \times 224 \times 416$, and $21 \times 112 \times 208$, respectively, see~\cref{fig:fig1}.
We finetune the model using the pyramidal flow matching loss~\cite{jin_pyramidal_2025}, demonstrating that this approach substantially reduces inference cost while maintaining near-original quality.
Furthermore, we conduct a study of various step distillation strategies within the pyramidal setup, both for conventional and pyramidal teacher diffusion models.
We also demonstrate for the first time that recently proposed Pyramidal Patchification models (an alternative to PyramidalFlow)~\cite{li2025pyramidalpatchificationflowvisual} can be successfully trained for few-step video generation.

In addition to this empirical study, we present a theoretical generalization of the resolution transition operations introduced in PyramidalFlow. 
Specifically, we extend these operations to arbitrary upsampling and downsampling functions based on orthogonal transforms. 
Notably, average pooling and nearest-neighbor upsampling, employed in the original work, can be interpreted as scaled instances of the Haar wavelet operator, fitting within our generalized framework.

In summary, our contributions are as follows:

\begin{enumerate}
\item We show that a conventional video diffusion transformer can be effectively converted into a spatiotemporal pyramidal diffusion model with minimal finetuning cost and without compromising quality.
\item We conduct a systematic study of step distillation techniques within the pyramidal setup, offering practical insights for various training scenarios.
\item We extend the procedure of transition between stages in the PyramidalFlow framework to a broader class of upsampling functions.

\end{enumerate}

%% file: tables/flops.tex
\begin{tabular}{llr}
\toprule
Inference method & Schedule & TFLOPs~$\downarrow$\\%
\midrule
Diffusion & 0-0-50 & 2\ $\times$\ \num{12592} \\%
Pyramidal diffusion & 20-20-10 & 2\ $\times$\ \num{2821} \\%
\midrule
\multirow{2}{*}{Step distillation} & 0-0-4 & \num{1007} \\%
& 0-0-2 & \num{504} \\%
\midrule
\multirow{3}{*}{Pyramidal step distillation} & 2-2-2 & \num{534} \\%
& 2-2-1 & \num{282} \\%
& 1-1-1 & \num{267} \\%
\bottomrule
\end{tabular}

%% file: tables/latency.tex
\begin{tabular}{clr}
\toprule
Method & Stage & Latency, ms~$\downarrow$\\%
\midrule
\multirow{3}{*}{\makecell{PyramidalWan\\$81 \times 448 \times 832$}} & 0 (hi-res) &  \num{631.77} \\%
 & 1 (mid-res) &  \num{33.76} \\%
 & 2 (lo-res) &  \num{7.62} \\%
\midrule
\multirow{3}{*}{\makecell{Wan-PPF \\$81 \times 480 \times 832$}} & 0 (hi-res) & 713.76 \\
& 1 (mid-res) & \num{39.85} \\
& 2 (lo-res) & \num{8.26}\\
\bottomrule
\end{tabular}

%% file: sec/2_related.tex
\section{Related works}
\label{sec:related_works}

\paragraph{Pyramidal models}
The observation that Gaussian noise degrades information across all frequency components at a uniform rate has been well established in the literature~\cite{dieleman2024spectral,falck2025fourierspaceperspectivediffusion}. 
Simultaneously, natural signals such as images and videos are known to exhibit relatively low magnitudes in their high-frequency components~\cite{hoogeboom_simple_2023,Hoogeboom_2025_CVPR}. 
Together, these insights motivate a multi-resolution denoising strategy.
An early implementation of this idea was the cascaded diffusion model, which employed multiple denoising networks, each operating at a distinct resolution~\cite{ho_cascaded_2021}. 
Subsequent works aimed to unify this approach within a single network, introducing different mechanisms for transitioning between resolutions~\cite{campbell2023transdimensional,chen2025pixelflow,zhang2025training}. 
PyramidalFlow~\cite{jin_pyramidal_2025} proposed a mathematically grounded framework based on flow matching, offering a coherent view of the forward diffusion process and spatial resolution changes. 
Building on this, TPDiff extended the methodology to the video frame rate varying per stage~\cite{ran2025tpdifftemporalpyramidvideo}. 
The practical viability of pyramidal models has also been demonstrated through deployment on resource-constrained devices such as mobile chips~\cite{anonymous2025neodragon}.
In our work, we adopt the PyramidalFlow framework to convert a pretrained video model into a pyramidal pipeline, enabling efficient inference with relatively cheap training.

\paragraph{Patch-pyramidal models}
An alternative to modifying input resolution is adjusting the kernel size of the patchification and unpatchification layers in the diffusion transformer (DiT) based on the noise level. 
This allows the most computationally intensive transformer blocks to operate on fewer tokens for noisier inputs, achieving similar efficiency gains to PyramidalFlow. 
An advantage of this approach is that it avoids the need for mathematical derivations to handle stage transitions.
FlexiDiT~\cite{Anagnostidis_2025_CVPR} combines this strategy with learnable per-stage LoRA adapters~\cite{hu2022lora}, which may pose challenges for inference on resource-constrained devices. More recently, Pyramidal Patchification Flow (PPF)~\cite{li2025pyramidalpatchificationflowvisual} demonstrated that such adapters are not essential, neither for training from scratch nor for finetuning pretrained text-to-image diffusion models.
In our experiments, we find that under limited training budgets, PyramidalFlow outperforms PPF for diffusion-style finetuning. 
Nevertheless, we show that patch-pyramidal models remain a strong candidate for distillation into few-step inference models.

\paragraph{Step distillation of diffusion models}
Since multi-step inference in diffusion models is computationally intensive for many applications, step distillation has become a key area of research. 
The most widely adopted methods include adversarial distillation~\cite{sauer2024fasthighresolutionimagesynthesis}, distribution matching distillation (DMD)~\cite{luo_diff-instruct_2023,yin_one-step_2024,yin_improved_2024}, and consistency models~\cite{chen_sana-sprint_2025,geng_mean_2025,sabour2025align}. 
While most efforts in this field have focused on image generation, several recent papers have extended these techniques to video models~\cite{pmlr-v267-lin25m,zheng_large_2025,Wu_2025_CVPR,Ben_Yahia_2025_ICCV}.
In our work, we explore DMD and adversarial approaches to distill a few-step pyramidal version of a pretrained video diffusion model, whether originally pyramidal or not. 
Notably, the concurrent SwD~\cite{anonymous2025scalewise} and Neodragon~\cite{anonymous2025neodragon} works also studied pyramidal step distillation to reduce inference costs.
However, SwD did not consider the case of a pyramidal teacher model, while Neodragon has not explored PPF-based training.
Our work fills these gaps.

%% file: sec/3_method.tex
\section{Preliminaries}
\label{sec:method}
\input{figures/method_overview}


  
    

To reduce the computational cost of video generation inference, we adopt the PyramidalFlow framework~\cite{jin_pyramidal_2025}, splitting the forward and backward diffusion processes into $S$ stages indexed by $i$. 
Each stage corresponds to a specific spatiotemporal resolution, where stage $i = 0$ operates at the original video tensor size, and stage $i = S - 1$ at the lowest resolution. 
In practice, we use $S = 3$.
The upsampling operation $\upsample$ transitions from stage $i + 1$ to $i$ by doubling the number of frames, height, and width using nearest-neighbor upsampling.
Its counterpart, $\downsample$, is implemented via 3D average pooling.
Unlike prior works of \citet{jin_pyramidal_2025} and \citet{ran2025tpdifftemporalpyramidvideo}, which applied stage-wise generation to either spatial or temporal dimensions, we apply $\upsample$ and $\downsample$ across all three video axes.

\subsection{Stage-wise definition of clean signal}
\label{sec:define_clean_signal}
Most recent video generation models operate in the latent space of a pretrained autoencoder (VAE). 
In this setting, the target signal $\x_0$ is defined as the output of the VAE encoder $\mathcal{E}$ applied to an input RGB video $\mathcal{V}$, \ie, $\x_0 = \mathcal{E}\lft(\mathcal{V}\rgt)$.
Therefore, there are two options for constructing the clean signal $\x_0^{\left(i\right)}$ at stage $i > 0$.
The first approach, adopted by \citet{jin_pyramidal_2025}, defines it as a latent signal, downsampled several times:
\begin{equation}
    \x_0^{\left(i\right)} = \downsample\circ\dots\circ\downsample\lft(\x_0\rgt). \tag*{\oneincircle}
\label{eq:vae_then_down}
\end{equation}
The second option performs downsampling  in the original RGB space before encoding with VAE,
\begin{equation}
    \x_0^{\left(i\right)} = \mathcal{E}\lft(\texttt{Down}\lft(\mathcal{V}, i\rgt)\rgt), \tag*{\twoincircle}
\label{eq:down_then_vae}
\end{equation} 
where $\texttt{Down}$ denotes a suitable resizing operation, \eg trilinear interpolation~\cite{anonymous2025scalewise}.
As we show below, the choice between these definitions depends on the specific training setup.

\subsection{Stage-wise forward process}
\label{sec:forward}
For each stage $i$, we define two boundary noise levels: a \underline{c}leaner level $\globlevel[i]_c$ and a \underline{n}oisier level $\globlevel[i]_n$, such that $0 \leq \globlevel[i]_c < \globlevel[i]_n \leq 1$. 
Here, $\globlevel = 0$ corresponds to a clean signal, and $\globlevel = 1$ to \iid Gaussian noise.
Given the clean signal $\x_0^{\upi}$ at the appropriate resolution, we compute two boundary samples using a shared noise tensor $\epsilon \sim \mathcal{N}\lft(0, I\rgt)$:
\begin{align}
    \y_c^{\upi} &= \left(1 - \globlevel[i]_c\right) \x_0^{\left(i\right)} &+ \globlevel[i]_c \epsilon, \label{eq:y_c_i} \\
    \y_n^{\upi} &= \left(1 - \globlevel[i]_n\right) \upsample\left(\x_0^{\upiplus}\right) &+ \globlevel[i]_n \epsilon. \label{eq:y_n_i} 
\end{align}
For the noise level $\globlevel$ such that $\globlevel[i]_c < \globlevel \leq \globlevel[i]_n$, the noised signal $\x_{\globlevel}^{\upi} $ is defined via linear interpolation between boundary samples
\begin{align}
    \x_{\globlevel}^{\left(i\right)} 
    &= \left(1-\loclevel\right) \y_c^{\upi} + \loclevel \y_n^{\upi} \\
    &= \left(1-\loclevel\right) \left(1 - \globlevel[i]_c\right) \x_0^{\left(i\right)} \nonumber \\ 
    &\qquad + \loclevel \left(1 - \globlevel[i]_n\right) \upsample\left(\x_0^{\upiplus}\right) \nonumber \\ 
    &\qquad + \globlevel \epsilon,
    \label{eq:forward}
\end{align}
where $\loclevel = \frac{\globlevel - \globlevel[i]_c}{\globlevel[i]_n - \globlevel[i]_c} $ is the \emph{local} noise level, $0 < \rho \leq 1.$ 
For clarity, in contrast to local level $\rho$ we refer to $\globlevel$ as the \emph{global} noise level, since it determines the total amount of added noise, as evident from~\cref{eq:forward}.
At the target stage $i=0$ global noise levels are also called \emph{natural} noise levels and denoted by $\snrlevel$.
Although, strictly speaking \cref{eq:forward} describes a  generalized stochastic interpolant~\cite{albergo2025stochasticinterpolantsunifyingframework} rather than diffusion, we will still use the term `diffusion' for simplicity.

Importantly, in PyramidalFlow framework, resampling operations $\upsample$ and $\downsample$, stage-wise clean signals,  and boundary noise levels are defined to satisfy the following equivalence of probabilistic distributions
\begin{equation}
    \upsamplegauss\left(\y_c^{\left(i+1\right)}\right) \stackrel{d}{=} \y_n^{\left( i\right)},
\label{eq:upsample_law}
\end{equation}
where $\stackrel{d}{=}$ means equality in distribution, and $\upsamplegauss$ is the upsampling operation $\upsample$ followed by the addition of some amount of non-\iid noise.
This noise aims to decorrelate the adjacent pixels after upsampling, and its parameters have been derived by \citet{jin_pyramidal_2025}.
In Supplementary, we generalize the functions $\downsample$, $\upsample$, and $\upsamplegauss$ from simple resampling operations (average pooling and nearest neighbor interpolation) to any resizing methods based on orthogonal transforms, \eg wavelets.
Our generalization involves sampling the missing high-frequency components before upscaling from Gaussian noise.
This enables decorrelation  even when the upscaling operation involves interaction between pixels.

\Cref{eq:upsample_law} implies that the `cleaner' boundary sample from stage $i+1$ after upsampling with $\upsamplegauss$ becomes a valid `noisier' boundary sample of stage $i$.
This establishes a relation between noise levels across two consecutive stages.
This relation, if applied recursively, allows to map any global noise level $\globlevel[i]$ at stage $i$ to a corresponding natural noise level $\snrlevel$ at stage 0.
In Supplementary, we show that natural noise levels corresponding to different stages do not overlap.
This ensures that the natural level $\snrlevel$ is  a unique conditioning value for the denoising network in the pyramidal setup, independent of the number of stages $S$.

The `noisier' bounds $\{\globlevel[i]_n\}_i$ are selected in practice via spectral analysis of noised samples at each stage.
Specifically, $\globlevel[i]_n$ should be large enough that high-frequency components become indistinguishable from  scaled Gaussian noise, \ie $p\lft(\y_n^{\left( i\right)} \mid \x_0^{\left(i\right)}\rgt) \approx \mathcal{N}\lft(\left(1-\globlevel[i]_n \right) \x_0^{\left(i\right)}, \, \left(\globlevel[i]_n\right)^2 I\rgt).$
This allows to downsample and proceed with the forward process at stage $i+1$ without loss of information~\cite{dieleman2024spectral,anonymous2025scalewise}.

\input{figures/qual_res}

\section{Method}
\subsection{Pyramidal finetuning}
\label{sec:pyr_finetuning}
To convert a pretrained conventional diffusion model to its pyramidal version, we apply finetuning with the dedicated loss function.

\paragraph{Flow matching loss}
We start with the pretrained open-sourced Wan2.1-1.3B model $F$.
For brevity, hereinafter we omit the conditioning text prompt $c$ in the notation.
Since this model was trained with flow matching loss~\cite{wan2025wanopenadvancedlargescale}, it approximates the derivative of the noised signal \wrt the noise level~\cite{lipman2023flow}, 
\begin{equation}
    F\lft(\x_{\globlevel}, \globlevel\rgt) \approx
    \mathbb{E} \left[\frac{d \x_{\globlevel}}{d\globlevel} \mid \x_{\globlevel} \right].
    \label{eq:learned_fow}
\end{equation}
To preserve this property during pyramidal finetuning, at each stage $i$  we define the objective for the student network $F_\theta$ as the derivative \wrt the global noise level, \ie
\begin{equation*}
    \frac{d \x_{\globlevel}^{\left(i\right)}}{d \globlevel}
    = \frac{d \x_{\globlevel}^{\left(i\right)}}{d \loclevel} \frac{d \loclevel}{d \globlevel} 
    = \frac{\y_n^{\upi} - \y_c^{\upi}}{\globlevel[i]_n - \globlevel[i]_c}.
\end{equation*}
The pyramidal loss is defined then as 
\begin{equation}
    L_{\textrm{pyr}}\left(\theta\right) = 
    \sum_i \mathbb{E}_{x_{0}^{\left(i\right)}} \mathbb{E}_{\epsilon} \mathbb{E}_{\loclevel} \left\lVert F_\theta \lft(\x_{\globlevel}^{\left(i\right)}, \snrlevel\rgt) - \frac{d \x_{\globlevel}^{\left(i\right)}}{d \globlevel} \right\rVert^2,
    \label{eq:pyramidal_flow_matching_loss}
\end{equation}
for uniformly distributed  local level $\loclevel \sim \textrm{Uni}\left(0, 1\right).$

\paragraph{Distillation loss}
In addition to flow matching, we apply a distillation loss to align student's partially denoised latents with the teacher's predictions.
For stage $i$ and global noise level $\globlevel$, $\globlevel[i]_c < \globlevel \leq \globlevel[i]_n$, we first map $\globlevel$ to the natural noise level $\snrlevel$ as discussed in~\cref{sec:forward}.
We sampled the high-resolution noise $\epsilon^{\left(0\right)}$  at stage 0, and construct the teacher’s prediction for the noisy input $\x_{\snrlevel} = \left(1 - \snrlevel\right) \x_0 + \snrlevel \epsilon^{\left(0\right)} $  as 
\begin{equation}
    \Tilde{\x}_{\snrlevel[i]_c} = \x_{\snrlevel} - \left(\snrlevel - \snrlevel[i]_c\right) \cdot F\lft(\x_{\snrlevel},\, \snrlevel\rgt),
\end{equation}
where $\snrlevel[i]_c$ is the natural noise level corresponding to the `cleaner' bound of the denoising process at stage $i$.

Next, we downsample the noise to stage $i$ and scale by a constant to preserve unit variance, $\epsilon^{\left(i\right)} \propto \downsample \circ \ldots \circ \downsample\lft(\epsilon^{\left(0\right)}\rgt).$
Using  $\epsilon^{\left(i\right)}$, we construct boundary samples $\y_c^{\upi}$ and $\y_n^{\upi}$  for the student's noised input $\x_{\globlevel}^{\upi}$.
Student's single-step prediction of the cleaner boundary value equals 
\begin{equation}
    \Tilde{\y}^{\upi}_c = \x_{\globlevel}^{\upi} - \left(\globlevel - \globlevel[i]_c\right) \cdot F_\theta\lft(\x_{\globlevel}^{\upi},\, \snrlevel\rgt).
\end{equation}
The distillation loss is then defined as
\begin{equation}
    L_\textrm{dist}\lft(\theta\rgt) = \sum_i \mathbb{E}_{x_{0}^{\left(i\right)}, \x_0} \mathbb{E}_{\epsilon^{\left(0\right)}} \mathbb{E}_{\globlevel} \left\lVert \Tilde{\y}^{\upi}_c - \Tilde{\x}_{\snrlevel[i]_c}  \right\rVert^2.
\end{equation}
In early experiments, we found that training with latents downsampled  in pixel space, \eg using Definition~\ref{eq:down_then_vae}, yields significantly better visual results.
This observation aligns with  findings of \citet{anonymous2025scalewise}.
We refer to the resulting model $F_\theta$ as \emph{PyramidalWan}.

\subsection{Pyramidal step distillation}
\label{sec:pyr_step_distill}
While pyramidal diffusion alone reduces computational cost by 78\% (see~\cref{tab:flops}), we aim to further decrease latency through step distillation. 
Below, we describe how distribution matching distillation (DMD) and adversarial technique are adapted to the pyramidal framework.

\subsubsection{DMD with original teacher}
\label{sec:pyr_dmd_original_teacher}
Since in case of Wan2.1 model we have access to the original pretrained non-pyramidal model $F$, we can use it as a teacher in a DMD pipeline~\cite{yin_improved_2024}.
In this method,  the student model $F_\xi$ is trained to predict the clean signal at the $i$-th stage in a single step, $\hat{\x}_0^{\upi} = \x_{\globlevel[i]} - \globlevel[i] \cdot F_\xi \lft(\x_{\globlevel[i]}, \snrlevel\rgt)$.
Notably, to construct the input $\x_{\globlevel[i]}$ we follow the rollout strategy proposed by \citet{yin_improved_2024}, using a detached output of the student:
\begin{equation}
    \x_{\globlevel[i]} = \y_n^{\upi} + \left(\globlevel[i] - \globlevel[i]_n\right) \cdot \texttt{stopgrad}\left[ F_\xi \lft(\y_n^{\upi}, \snrlevel_n^{\upi}\rgt) \right].
\end{equation}
Once the clean signal $\hat{\x}_0^{\upi}$  is predicted, it is re-noised according to the  teacher's forward process  with a noise level $\globlevel'$, $0 < \globlevel' \leq 1$,
 \begin{equation*}
    \hat{\x}_{\globlevel'}^{\left(i\right)} = \left(1 - \globlevel'\right) \hat{\x}_{0}^{\left(i\right)} + \globlevel' \varepsilon.
\end{equation*}
The \emph{fake score} network $F_\varphi$ is trained using regular flow matching loss, aiming to denoise the student's predictions,
\begin{equation*}
    L_{\textrm{fm}}\lft(\varphi\rgt) = \sum_i\mathbb{E}_{\hat{\x}_{\globlevel'}^{\left(i\right)}} \left\lVert F_\varphi\lft(\hat{\x}_{\globlevel'}^{\left(i\right)}, \globlevel'\rgt) - \dfrac{d \hat{\x}_{\globlevel'}^{\left(i\right)}}{d \globlevel'} \right\rVert^2.
\end{equation*}
The gradient of DMD loss is defined as follows,
\begin{align}
    & \nabla_\xi L_{\textrm{dmd}}\lft(\xi\rgt) = \nonumber \\  
    &\quad 
    \left(F_\varphi \lft(\hat{\x}_{\globlevel'}^{\left(i\right)}, \globlevel'\rgt) - F\lft(\hat{\x}_{\globlevel'}^{\left(i\right)}, \globlevel'\rgt) \right) \cdot \nabla_\xi F_\xi \lft(\x_{\globlevel[i]}, \snrlevel\rgt).
    \label{eq:dmd_ot_loss}
\end{align}
For each training sample, weight of the loss $ L_{\textrm{dmd}}$ is set equal to $w_{\textrm{dmd}}$,
\begin{align}
    w_{\textrm{dmd}} &= \globlevel[i] \cdot\left\lVert F \lft(\hat{\x}_{\globlevel'}^{\left(i\right)}, \globlevel'\rgt) - \dfrac{d \hat{\x}_{\globlevel'}^{\left(i\right)}}{d \globlevel'} \right\rVert_1^{-1},
\end{align}
giving greater weight to samples whose re-noised versions can be well `denoised' by the teacher $F$.
To stabilize training, we also include a supervised loss that encourages the student's output to be similar to the teacher's one~\cite{sabour2025align}, $L_{\textrm{teach}} \lft(\xi\rgt) = \left\lVert F_\xi \lft(\x_{\globlevel[i]}, \snrlevel\rgt) - F\lft(\x_{\globlevel[i]}, \globlevel[i]\rgt) \right\rVert^2$.
This term is added to the DMD loss with a weight of \num{0.01}.

We found experimentally that such training does not work for the original pretrained Wan2.1-1.3B teacher due to its inability to generate videos at the lowest spatiotemporal resolution, \ie for $i=S-1$.
While this may seem to contradict the results recently reported by~\citet{anonymous2025scalewise}, we note that their SwD method gradually upscales the video tensor by a fractional factor after each step, likely reaching the resolution compatible with the teacher model earlier in the generation process.
In our case, to remain consistent with PyramidalFlow framework, we briefly finetuned the teacher using flow matching loss on a dataset of videos with varying resolution. 
Clean latent tensors  were generated using Definition~\ref{eq:down_then_vae}, which was also used for training.
In the beginning of step distillation, both the student and fake score network were initialized  from the finetuned teacher's checkpoint.
At inference time, we use the same sampling algorithm as in pyramidal diffusion, but with only a few steps per stage~\cite{Lin_2024_WACV}.
The resulting few-step generator is referred to as \emph{PyramidalWan-DMD-OT}.

\vspace{-0.25em}

\subsubsection{DMD with pyramidal teacher}
\label{sec:dmd_pyr_teacher}
The DMD pipeline described above requires modifications when employing a pyramidal flow matching model as the teacher.
This setup is increasingly relevant given the reported training efficiency of pyramidal diffusion models~\cite{li2025pyramidalpatchificationflowvisual,minimax_team_minimax_2025}.
The key difference arises from the fact that pyramidal teacher has been trained with stage-wise inputs.
Therefore, student's prediction of the clean signal $\hat{\x}_0^{\upi}$  should be re-noised similarly to \cref{eq:y_n_i,eq:y_c_i},
\begin{align}
    \hat{\y}_c^{\upi} &= \left(1 - \globlevel[i]_c\right) \hat{\x}_0^{\left(i\right)} &+ \globlevel[i]_c \varepsilon, \\
    \hat{\y}_n^{\upi} &= \left(1 - \globlevel[i]_n\right) \upsample\circ\downsample\left(\hat{\x}_0^{\left(i\right)}\right) &+ \globlevel[i]_n \varepsilon, \label{eq:y_n_i_renoise} \\
    \hat{\x}_{\globlevel'}^{\left(i\right)} 
    &= \left(1-{\loclevel}'\right) \hat{\y}_c^{\upi} + {\loclevel}' \hat{\y}_n^{\upi}.
    \label{eq:renoising_dmd_pyr}
\end{align}
This definition of $\hat{\y}_n^{\upi}$ assumes that per-stage clean signals follow Definition~\ref{eq:vae_then_down}, in contrast to how the the pyramidal diffusion model has been trained (see~\cref{sec:pyr_finetuning}). 
Using Definition~\ref{eq:down_then_vae} would require  both VAE decoder and encoder to compute $\hat{\x}_0^{\upiplus}$, which is computationally expensive for video models.
To address this, before applying the DMD pipeline, we finetuned our  pyramidal diffusion model $F_\theta$ according to this definition, resulting in the model $F_{\theta_1}$.
We found that this approach works better than training with Definition~\ref{eq:vae_then_down} from the beginning.
Importantly, using the original $F_\theta$ as a teacher leads to unsatisfactory quality of few-step generations.

The fake score network $F_\varphi$ is trained with pyramidal flow matching loss $L_{\textrm{pyr}}\lft(\varphi\rgt)$ from \cref{eq:pyramidal_flow_matching_loss}, but with re-noised predicted signals $\hat{\x}_0^{\upi}$ instead of ground-true noised signals $\x_0^{\upi}$.

The original DMD formulation relies on estimating  the score function, or equivalently, the added Gaussian noise $\varepsilon$~\cite{luo_diff-instruct_2023,yin_one-step_2024,albergo2025stochasticinterpolantsunifyingframework}.  
To construct such an estimator, we define $\Delta^{\upi}$, a linear combination of $\hat{\y}_c^{\upi}$ and $\hat{\y}_n^{\upi}$,
\begin{align}
    \Delta^{\upi} &= \globlevel[i]_n \hat{\y}_c^{\upi} - \globlevel[i]_c \hat{\y}_n^{\upi} \\
    &= \globlevel[i]_n \left(1 - \globlevel[i]_c\right)  \hat{\x}_0^{\upi} \nonumber \\
    &\qquad - \globlevel[i]_c \left(1 - \globlevel[i]_n\right)  \upsample\circ\downsample\lft( \hat{\x}_0^{\upi}\rgt).
\end{align}
Applying $\upsample\circ\downsample$ to both LHS and RHS and using the fact that $\upsample\circ\downsample\circ\upsample\circ\downsample = \upsample\circ\downsample$, gives
\begin{equation}
    \upsample\circ\downsample\lft(\Delta^{\upi}\rgt) = \left(\globlevel[i]_n - \globlevel[i]_c\right) \upsample\circ\downsample\lft(\hat{\x}_0^{\upi}\rgt).
\end{equation}
Together with \cref{eq:y_n_i_renoise}, this yields a  closed-form expression for $\varepsilon$ given $\hat{\y}_n^{\upi}$ and $\hat{\y}_c^{\upi}$.

For a pretrained pyramidal teacher or pyramidal fake score model $F_\nu$, where $\nu \in \left\{ \theta_1, \varphi \right\}$, the stage boundary samples are estimated as $\hat{\x}_{\globlevel'}^{\upi} + \left(\globlevel[i]_n - \globlevel'\right) F_\nu \lft(\hat{\x}_{\globlevel'}^{\upi}, \snrlevel'\rgt)$ and $\hat{\x}_{\globlevel'}^{\upi} + \left(\globlevel[i]_c - \globlevel'\right) F_\nu \lft(\hat{\x}_{\globlevel'}^{\upi}, \snrlevel'\rgt)$ respectively.

The noise estimator $\hat{\varepsilon}_\nu$  equals then (we omit arguments of $F_\nu$ for brevity)
\begin{align*}
    \hat{\varepsilon}_\nu &=
    \frac{1}{\globlevel[i]_n} \left(\hat{\x}_{\globlevel'}^{\upi} + \left(\globlevel[i]_n - \globlevel'\right) F_\nu \right) \nonumber \\
    &\quad - \frac{1 - \globlevel[i]_n}{\globlevel[i]_n}\left(\upsample\circ\downsample\lft(\hat{\x}_{\globlevel'}^{\upi}\rgt) - \globlevel' \cdot \upsample\circ\downsample\lft(F_\nu\rgt) \right).
\end{align*}
The difference between the estimators provided  by the teacher and the fake score model is proportional to 
\begin{align*}
   \hat{\varepsilon}_\varphi - \hat{\varepsilon}_{\theta_1} &\propto 
   \beta_1 \cdot \left(F_\varphi - F_{\theta_1}\right) + \beta_2 \cdot \upsample\circ\downsample\lft(F_\varphi - F_{\theta_1}\rgt),
\end{align*}
where $
    \beta_1 = \globlevel[i]_n - \globlevel', \
    \beta_2 = \globlevel' \left(1 - \globlevel[i]_n \right).
$
We normalize these weights as $\tilde{\beta}_k = \frac{\beta_k}{\beta_1 + \beta_2}$ for $k=1,2.$
Similarly, the gradient of the re-noised prediction is  proportional to the following weighted sum
\begin{align}
    - \nabla_\xi \hat{\x}_{\globlevel'}^{\upi}  \propto
    \gamma_1 \nabla_\xi F_\xi + \gamma_2 \nabla_\xi \upsample\circ\downsample\lft(F_\xi\rgt),
\end{align}
with the weights $
    \gamma_1 = \left(\globlevel[i]_n - \globlevel'\right)\left(1 - \globlevel[i]_c\right),
    \gamma_2 = \left(\globlevel' - \globlevel[i]_c\right)\left(1 - \globlevel[i]_n\right).
$
We normalize these  as $\tilde{\gamma}_k = \frac{\gamma_k}{\gamma_1 + \gamma_2}$ and define the gradient of pyramidal DMD loss as 
\begin{align}
    &\nabla_\xi L_{\textrm{dmd-pyr}}\lft(\xi\rgt) = \left(\tilde{\beta}_1 \left(F_\varphi - F_{\theta_1}\right) + \tilde{\beta}_2 \upsample\circ\downsample\lft(F_\varphi - F_{\theta_1}\rgt)\right) \nonumber \\
    &\quad\quad \cdot \nabla_\xi \left( \tilde{\gamma}_1 F_\xi + \tilde{\gamma}_2 \upsample\circ\downsample\lft(F_\xi\rgt) \right).
    \label{eq:dmd_pyr_loss}
\end{align}
Each sample in  $L_{\textrm{dmd-pyr}}$ loss is weighted with \begin{equation}
    w_{\textrm{dmd-pyr}} = \loclevel \cdot\left\lVert F_{\theta_1} \lft(\hat{\x}_{\globlevel'}^{\left(i\right)}, \snrlevel'\rgt) - \dfrac{d \hat{\x}_{\globlevel'}^{\left(i\right)}}{d \globlevel'} \right\rVert_1^{-1}.
\end{equation}
We call the resulting model \emph{PyramidalWan-DMD-PT}.

In addition, we explored the simplified version by setting $\tilde{\beta}_1 = \tilde{\gamma}_1 = 1$ and $\tilde{\beta}_2 = \tilde{\gamma}_2 = 0$ in~\cref{eq:dmd_pyr_loss}.
This reduces  $L_{\textrm{dmd-pyr}}$ to the formulation similar to~\cref{eq:dmd_ot_loss}.
Although such  a variant has insufficient theoretical grounding, we found that it performs marginally better in practice.
This modification is referred to as \emph{-PT}$^{*}$.

\subsubsection{Adversarial distillation}
\label{sec:pyr_adv_dist}
As an alternative to DMD, we explore pyramidal adversarial distillation.
In this setting, student $F_\xi$ similarly predicts a `cleaner' boundary sample $\hat{\y}_c^{\upi}$  of the current stage $i$ in a single step, while a discriminator attempts to distinguish between features extracted from generated and  ground-true samples. 
The discriminator consists of two components:
\begin{enumerate}
    \item Frozen feature extractor $F^\dagger$, based on a pretrained diffusion model. We denote the model as \emph{PyramidalWan-Adv-OD} when using the backbone from the \underline{o}riginal Wan model $F$, and \emph{-PD} when using the  \underline{p}yramidal backbone of $F_\theta$ instead.
    \item Trainable discriminator head $D_\varphi$ attached to the final block of $F^\dagger$, comprises \textit{spatial} and \textit{temporal} branches implemented with lightweight convolutional layers and residual blocks.
\end{enumerate}
The discriminator minimizes an adversarial Hinge loss~\cite{lim2017geometricgan} over features at each stage,
\begin{equation*}
    \begin{aligned}
        L_\textrm{D}\lft(\varphi\rgt) = \sum_i \Big[ &
        \mathbb{E}_{\y^{\upi}_c} \big[ \max(0, 1 - D_\varphi(F^\dagger(\y^{\upi}_c))) \big] \\
        &+ \mathbb{E}_{\hat{\y}^{\upi}_c } \big[ \max(0, 1 + D_\varphi(F^\dagger(\hat{\y}^{\upi}_c ))) \big]
        \Big].
    \end{aligned}
\end{equation*}
The student optimizes a combined objective balancing adversarial and reconstruction terms, weighted by $\lambda_{\text{adv}}$ and $\lambda_{\text{rec}}$, similarly to the approach of \citet{zhang2024sfv}.
Empirically, we find that $\lambda_{\text{adv}} = 1$ and $\lambda_{\text{rec}} = 2$ yield the highest visual quality.
{\small
    \begin{align*}
        L_\textrm{G}\lft(\xi\rgt) = \sum_i 
        \mathbb{E}\big[ &- \lambda_{\text{adv}} \cdot D_\varphi(F^\dagger(\hat{\y}^{\upi}_c)) \nonumber
        + \lambda_{\text{rec}} \cdot \| \hat{\y}^{\upi}_c  - \y^{\upi}_c  \|_2^2 \big] 
    \end{align*}
} 

\subsection{Patch-pyramidal training}
An alternative approach for varying per-stage computational cost is Pyramidal Patchification Flow (PPF)~\cite{li2025pyramidalpatchificationflowvisual}.
PPF does not alter the resolution of the denoiser transformer's inputs or outputs. 
Instead, it introduces stage-wise patchification and unpatchification layers, as \cref{fig:fig1} shows.
For earlier stages, kernel size of patchifier is accordingly increased, and therefore, the transformers blocks in PPF operate with exactly the same number of tokens as for PyramidalFlow.
Importantly, diffusion training, step distillation, and inference can be performed within the PPF framework in the same way as for the original pretrained Wan model.

%% file: figures/method_overview.tex
\begin{figure*}[t]
\centering
\includegraphics[width=2\columnwidth]{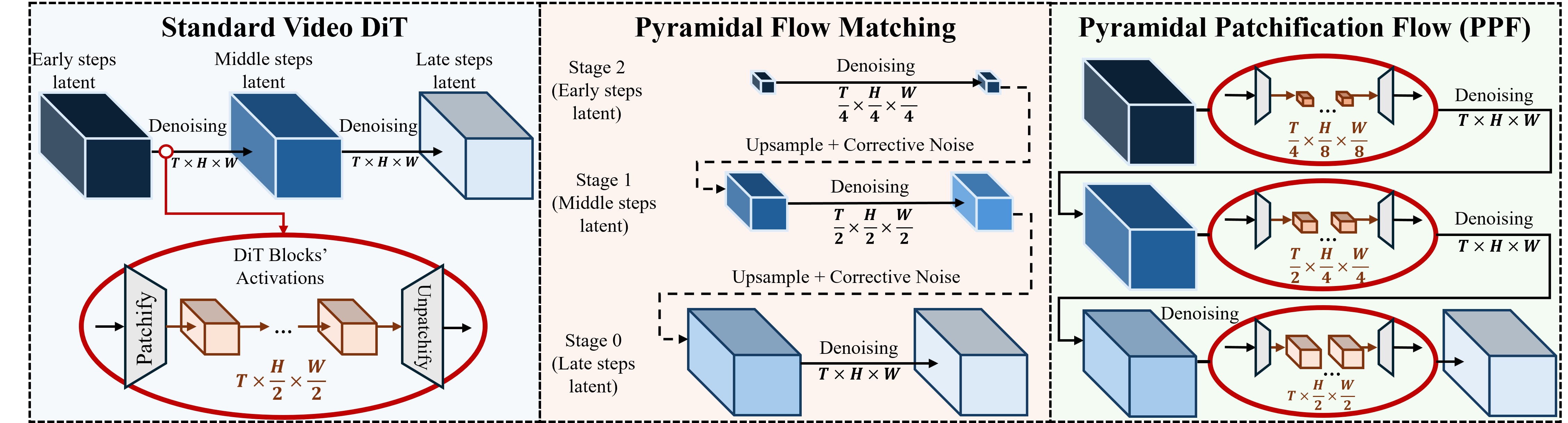}
\caption{\textbf{Inference of different types of models.}
Left: input and output tensors of a standard DiT always have the same size, and the number of tokens in transformer blocks does not depend on the noise level.
Center: in pyramidal flow matching, higher noise levels are processed at smaller spatiotemporal resolution. For transition between stages special corrective noise should be added after upsampling.
Right: in PPF framework instead of changing the resolution, kernel size of patchifier is adjusted for each stage. This keeps the number of tokens equal to that in pyramidal flow matching.
}
\label{fig:fig1}
\end{figure*}

%% file: figures/qual_res.tex
\begin{figure*}[t]
\centering
\includegraphics[width=2\columnwidth]{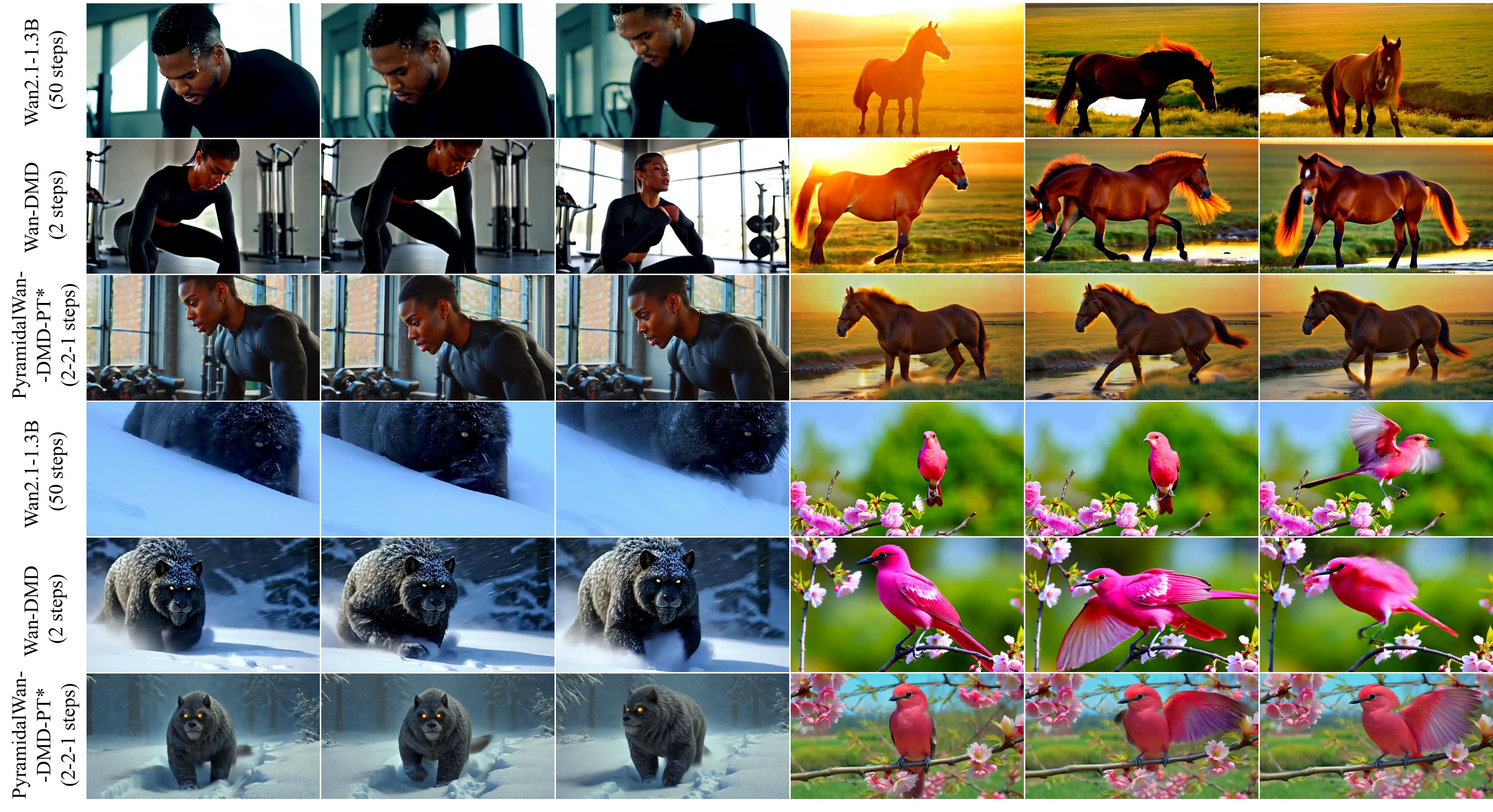}
\caption{\textbf{Examples of video generations.}
Videos produced by our pyramidal step-distilled model are similar in quality to outputs of more computationally expensive baselines.
}
\label{fig:qual_res}
\vspace{-1em}
\end{figure*}

%% file: sec/4_experiments.tex
\section{Experiments}
\label{sec:experiments}
\begin{table*}[t]
\centering
\caption{\textbf{Model comparison.} 
Step-distilled versions of the original model, Wan-DMD and Wan-Adv, provide the quality of generation in the few-step mode on par with the diffusion model. 
However, they cannot unlock the satisfactory single-step inference. 
Pyramidal models with a single step at highest resolution fill the gap and demonstrate good performance, as measured quantitatively.
}
\label{tab:vbench_aggregated}
\resizebox{\linewidth}{!}{%
\input{tables/vbench-agg}
}
\vspace{-1em}
\end{table*}

\subsection{Training setup}
\paragraph{Step distillation of the original model}
As a natural baseline for reducing inference cost, we trained a step-distilled version of the original Wan model. 
Following the DMD pipeline described in~\cref{sec:pyr_dmd_original_teacher}, we adopted the one-step rollout strategy~\cite{yin_improved_2024} but omitted the adversarial loss, as we found it unnecessary. 
This training setup does not require visual data and relies solely on text prompts; we used prompts from the 350K subset of the dataset provided by~\citet{lin2024opensoraplanopensourcelarge}. 
Training was conducted for 31K iterations on 16 H100 GPUs.
As an alternative, we performed adversarial distillation on the original Wan model, following the procedure similar to the one outlined in~\cref{sec:pyr_adv_dist}. 
The model was trained for 30K iterations on a single H100 GPU using 80K synthetic videos generated by Wan2.1-14B, a larger variant of the original model. 

\paragraph{Pyramidal flow matching and step distillation}
For these experiments, we used the same synthetic dataset of 80K videos generated by the Wan2.1-14B model.
We observed that models trained on the synthetic data produce visually superior results compared to those trained on real video samples.
To ensure compatibility with Wan’s patchification layer at the lowest stage ($i = 2$), we slightly reduced the spatial resolution of videos from the default $480 \times 832$ to $448 \times 832$, making thus both height and width divisible by 64.
All models mentioned in~\cref{sec:pyr_finetuning,sec:pyr_step_distill} were fine-tuned for 5K iterations with a batch size of 6 per GPU (2 samples per stage) on two H100 GPUs.
For DMD-PT and DMD-PT$^{*}$ we conducted training with LoRA adapters, since we found that otherwise training might diverge.
In other cases, we finetuned all the weights.

\paragraph{Patch-pyramidal training}
To train the diffusion-based Wan-PPF model, we used the same setup as for PyramidalWan. 
Training details of patch-pyramidal DMD model were kept consistent with those used for DMD distillation of the original Wan model, except for the smaller training budget. 
This experiment was conducted for 5K steps on 8 GPUs, with a batch size of 3 per GPU (one sample per stage). 
We did not observe improvements with longer training.

\subsection{Results}
\paragraph{Main results}
We evaluate the quality of generated videos using the VBench and VBench-2.0 toolkits~\cite{huang2024vbench,zheng2025vbench}.
Our results are summarized in~\cref{tab:vbench_aggregated}.
First, we observe that PyramidalWan, \ie the pyramidized diffusion model, achieves scores comparable to the original Wan model sampled with 50 steps, while being approximately \num{4.5} times more efficient in terms of FLOPs. 
Notably, it also achieves the highest Semantic score among all evaluated models, resembling the findings of the concurrent work of~\citet{zhang2025training}.

While \citet{li2025pyramidalpatchificationflowvisual} successfully demonstrated finetuning of pretrained text-to-image models within the PPF framework, we found that extending this approach to video generation is  challenging. 
Using the same training budget and dataset as in our pyramidal training setup, our PPF-based text-to-video diffusion model failed to converge, and the quality of generated clips remained unsatisfactory.
Notably, increasing the compute budget to 8 GPUs did not improve results.
Step distillation using DMD also failed when the new patchification and unpatchification layers were initialized according to the scheme proposed in the original PPF work. 
Surprisingly, however, the DMD pipeline could still be applied successfully when the student model was initialized with the pretrained PPF diffusion model --- even though that chekpoint itself produced poor-quality generations. 
We attribute this to the mode-seeking behavior of the reverse-KL objective used in DMD~\cite{xu_one-step_2025}.

Among few-step models, the step-distilled baseline Wan-DMD provides  very strong performance combined with efficient inference.
Despite the significant gains in test-time efficiency (see~\cref{tab:flops}), it  surpasses the original diffusion model in both total scores even when sampled with only two steps.
However, single-step generation  remains infeasible as the quality of videos drops substantially.
We fill this gap with our few-step pyramidal models.
They are evaluated in a scenario with only one step at highest resolution, \ie at stage $i=0$, and a few steps at lower-resolution stages.
Given the much lower computational cost and latency of these stages, as reported in~\cref{tab:flops,tab:latency}, we adopt a 2-2-1 inference schedule (steps per stage in resolution-increasing order).

All models under this schedule achieve total score of VBench comparable to Wan diffusion model, with only a minor drop relative to  Wan-DMD sampled with 2 steps.
VBench-2.0, however, indicates some degradation, in particular for Creativity and Controllability dimensions.
Although distillation with the original teacher, PyramidalWan-DMD-OT, gets the best metrics within this set of models, we observed that its outputs often exhibit oversaturated colors and  cartoon-alike appearance.
Please refer to Supplementary for visual examples.
Among all models, we found that PyramidalWan-DMD-PT$^{*}$ produced the most visually appealing results, and therefore selected it for the user study.

In the study, the assessors were shown pairs of videos generated from identical prompts and asked to choose the preferred one or select ``no preference''.
Each pair included one video generated with step-distilled pyramidal model, and  one from either Wan with 50 steps (first study) or  Wan-DMD with 2 steps (second).
In total, we collected 700 responses.
We conducted the binomial test for the hypothesis `Baseline is strictly preferred with probability 0.5' and the one-sided `less' alternative.
As shown in~\cref{tab:user_study}, in both comparisons the hypothesis should be rejected.
This indicates that the participants found the quality of this model  on par with more computationally expensive baselines despite its lower VBench-2.0 score.
Examples of videos are provided in~\cref{fig:qual_res}.

\begin{table}[t]
\centering
\caption{\textbf{User study.} 
We evaluate our pyramidal model with 2-2-1 inference schedule against two baselines.
As the results show, participants did not find significant difference in visual quality.
}
\label{tab:user_study}
\resizebox{0.85\linewidth}{!}{%
\input{tables/human_preference}
}
\caption{\textbf{Ablation study.} 
Training without distillation loss improves the VBench-2.0 score but reduces the amount of motion in generated videos.
Simplified version of DMD objective leads to better empirical results.
}
\label{tab:ablation}
\resizebox{0.85\linewidth}{!}{%
\input{tables/ablation}
}
\vspace{-2em}
\end{table}

\paragraph{Ablations}
To check the impact of distillation losses in both pyramidal finetuning and step distillation experiments, we conducted experiments by removing this terms.
For PyramidalWan, this reduced VBench-2.0 total score from 54.93 to 54.02.
Notably, for DMD with pyramidal teacher effect was the opposite (see~\cref{tab:ablation}), however  at the expense of noticeable reduction of Dynamic Degree.
As mentioned in~\cref{sec:dmd_pyr_teacher}, the simplified version of $L_\textrm{dmd-pyr}$ objective yields  improved scores despite its theoretical weakness.
We leave further investigation  of this phenomenon for the future.

%% file: tables/vbench-agg.tex
\begin{tabular}{lcccccccccc}
\toprule
\multirow{2}{*}{\textbf{Model}} & \multicolumn{3}{c}{\textbf{VBench~$\uparrow$}} && \multicolumn{6}{c}{\textbf{VBench-2.0~$\uparrow$}} \\
\cmidrule{2-4} \cmidrule{6-11}
 & \textbf{Total} & \textbf{Quality} & \textbf{Semantic} && 
 \textbf{Total} & \textbf{Creativity} & \textbf{Commonsense} & \textbf{Controllability} & \textbf{Human Fidelity} & \textbf{Physics} \\
\midrule
Wan2.1-1.3B (50 steps)%
& 82.49 & 83.47 & 78.57 %
&& 56.02 & 48.73 & 63.38 & 33.96 & 80.71 & 53.30 \\%
Wan2.1-1.3B (25 steps)%
& 80.87 & 82.09 & 76.02 %
&& 55.73 & 49.49 & 62.50 & 35.01 & 79.44 & 52.20 \\%
PyramidalWan (20-20-10) %
& 82.83 & 83.36 & 80.70 %
&& 54.93 & 44.64 & 64.33 & 28.39 & 85.38 & 51.89 \\%
\midrule
Wan-Adv (4 steps)%
& 82.72 & 84.06 & 77.39 %
&& 55.40 & 50.96 & 57.33 & 32.56 & 85.58 & 50.57\\%
Wan-Adv (2 steps)%
& 82.35 & 83.74 & 76.82 %
&& 54.82 & 47.07 & 58.47 & 31.51 & 82.72 & 54.36 \\%
Wan-Adv (1 step)%
& 80.28 & 81.38 & 75.85 %
&& 50.36 & 38.51 & 54.41 & 29.99 & 79.60 & 49.29 \\%
Wan-DMD (4 steps)%
& 83.33 & 84.71 & 77.86 %
&& 57.48 & 50.87 & 58.16 & 36.36 & 82.80 & 59.23\\%
Wan-DMD (2 steps)%
& 83.28 & 84.00 & 80.41 %
&& 56.67 & 47.00 & 62.81 & 37.07 & 80.73 & 55.72 \\%
Wan-DMD (1 step)%
& 79.45 & 80.63 & 74.75 %
&& 53.17 & 38.69 & 59.31 & 35.36 & 77.49 & 55.00 \\%
\midrule
PyramidalWan-Adv-OD (2-2-1)%
& 82.90 & 83.94 & 78.74 %
&& 52.29 & 44.80 & 60.86 & 22.86 & 81.74 & 51.17 \\%
PyramidalWan-Adv-PD (2-2-1)%
& 82.57 & 84.20 & 76.07 %
&& 54.30 & 45.00 & 61.98 & 25.85 & 87.65 & 51.01 \\%
Wan-PPF-DMD (2-2-1) %
& 82.39 & 83.04 & 79.80 %
&&  53.45 & 40.04 & 62.84 & 24.76 & 91.61 & 47.99 \\%
PyramidalWan-DMD-OT (2-2-1) %
& 82.86 & 83.63 & 79.80 %
&& 55.36 & 50.04 & 64.34 & 29.05 & 77.50 & 56.78 \\%
PyramidalWan-DMD-PT$^{*}$ (2-2-1) %
& 82.72 & 83.46 & 79.75 %
&& 51.75 & 34.81 & 63.18 & 28.48 & 81.38 & 50.92 \\%
\bottomrule
\end{tabular}

%% file: tables/human_preference.tex

\begin{tabular}{lcccc}
\toprule
\multirow{2}{*}{Baseline} & \multicolumn{3}{c}{Preference, \%~$\uparrow$} & \multirow{2}{*}{$p$-value} \\%
\cmidrule{2-4} 
& Ours & No preference & Baseline & \\%
\midrule
Wan (50 steps) & 29.1 & 29.1 & 41.7 & $<$ \num{0.001} \\%
Wan-DMD (2 steps) & 33.1 & 35.4 & 31.4 & $<$ \num{0.001} \\
\bottomrule
\end{tabular}

%% file: tables/ablation.tex
\begin{tabular}{lrr}
\toprule
\multirow{2}{*}{Model} & \multicolumn{2}{c}{Total score~$\uparrow$} \\%
\cmidrule{2-3} 
& VBench & VBench-2.0 \\%
\midrule
PyramidalWan-DMD-PT$^{*}$ & 82.72 & 51.75 \\%
PyramidalWan-DMD-PT$^{*}$ w/o $L_\textrm{teach}$ & 82.44 & 52.36 \\%
PyramidalWan-DMD-PT & 82.56 & 50.67 \\%
\bottomrule
\end{tabular}

%% file: sec/5_conclusion.tex
\section{Conclusion}
\label{sec:conclusion}

In this work we explored pyramidization --- a strategy of reducing inference costs of video diffusion models that is  complementary to other architectural innovations.
We presented a pipeline of converting a pretrained \emph{conventional} diffusion model into a pyramidal one, both for multi-step and few-step inference regimes.
Further, we demonstrated  step distillation of a pretrained \emph{pyramidal} diffusion model: an important milestone for future research given the reported training efficiency of such systems.
In addition, we made a theoretical contribution by extending the procedure of switching between resolutions to a broader class of upsampling operations.
The resulting models occupy the practical niche enabling few-step generation with only a single step at the target resolution.
While demonstrating results comparable to more costly baselines in the human preference study, our models still lag behind in certain quantitative metrics.
Addressing this gap remains a promising direction for future work.

%% file: supplementary/X_suppl.tex
\clearpage
\renewcommand{\thetable}{A\arabic{table}}
\renewcommand{\thefigure}{A\arabic{figure}}
\renewcommand{\thesection}{A\arabic{section}}
\renewcommand{\thepage}{A\arabic{page}}
\setcounter{table}{0}
\setcounter{page}{1}
\setcounter{section}{0}
\maketitlesupplementary

This supplementary material is structured as follows.
\cref{sec:generalization_avgpool} provides our theoretical contribution: generalization of transition between stages to a broader class of resizing operations.
In~\cref{sec:add_experiments} we report additional experimental results: we apply recent Jenga~\cite{zhang2025training} method to our Wan-DMD checkpoint and compare its latency with our pyramidal approach.
\cref{sec:details} contains training and evaluation details necessary for reproducing our research.
We provide the extended version of \cref{tab:vbench_aggregated} of the main text in~\cref{tab:vbench_full_aggregated,tab:vbench2_full_aggregated}.
We encourage the readers to review the attached videos and compare the outputs produced by different models qualitatively.


%% file: supplementary/wavelets.tex
\section{Transition between pyramidal stages}  
\label{sec:generalization_avgpool}
In this section we provide a generalization for nearest-neighbour upsampling and average pooling used for transition between stages in PyramidalFlow.
While we describe our apporach in terms of wavelets, note that the same derivations are valid for any resizing operation based on orthogonal transforms.

\subsection{Downsampling}
\label{sec:downsampling}
Consider a clean video tensor $\x_0$ at the desired output resolution. 
For the purpose of this section we treat it as a single-channel tensor, flattened in a column vector $\x_0 \in \mathbb{R}^{T \cdot H \cdot W}.$
We use a forward diffusion process and denote the noisy version of this video as $\x_{\globlevel} = \left(1 - \globlevel\right) \x_0 + \globlevel \epsilon$ for some independently sampled $\epsilon \sim \mathcal{N}\left(0, I\right),$ \ie $p\lft(\x_{\globlevel} \mid \x_0\rgt) = \mathcal{N}\left(\left(1 - \globlevel\right) \x_0, \globlevel^2 I \right).$

For a single-level orthogonal wavelet decomposition with analysis matrix $\mathcal{W}$ we can decompose $\x_{\globlevel}$ into low- and high-frequency bands $\mathcal{W}\x_{\globlevel} = \left(\hat{\x}_{\globlevel,\lo}^T, \, \hat{\x}_{\globlevel,\hi}^T\right)^T \in \mathbb{R}^{T \cdot H \cdot W},$ where $\hat{\x}_{\globlevel,\lo}^T$ has a dimension of $\frac{T}{2} \cdot \frac{H}{2} \cdot \frac{W}{2}.$ 
To ``extract'' the low-frequency part, we can use the projector matrix $\Pi_\lo$ and write $\hat{\x}_{\globlevel,\lo} = \Pi_\lo \mathcal{W} \x_{\globlevel}.$
Due to the properties of linear operators, the distribution of $p\lft(\hat{\x}_{\globlevel,\lo} \mid \x_0\rgt)$ is still Gaussian, and we can compute its mean vector and covariance matrix.
For covariance we have
\begin{align*}
    \textrm{Cov}\lft[\hat{\x}_{\globlevel,\lo} \mid \x_0\rgt] 
    &= \Pi_\lo \mathcal{W} \cdot \globlevel^2 I  \cdot \left(\Pi_\lo \mathcal{W} \right)^T \\
    &= \globlevel^2 \Pi_\lo \mathcal{W} \mathcal{W}^T \Pi_\lo^T \\
    &= \globlevel^2 \Pi_\lo \Pi_\lo^T \\ 
    &= \globlevel^2  I
\end{align*}
thanks to the orthogonality of matrix $\mathcal{W}.$
For the mean,
\begin{align*}
    \mathbb{E}\lft[\hat{\x}_{\globlevel,\lo} \mid \x_0\rgt] 
    &= \left(1 - \globlevel\right) \Pi_\lo \mathcal{W}  \x_0 = \left(1 - \globlevel\right) U \x_0,
\end{align*}
where $U$ is a low-frequency part of the matrix $\mathcal{W} = \left(U^T, \; V^T\right)^T.$

In general, row sums for entries of $U$ are not equal to 1.
Therefore, if $\x_0$ is a constant vector with all entries being equal to each other, pixel values in $U \x_0$ will differ from those in $\x_0$.
To compensate that, for the given  wavelet $\mathcal{W}$ we introduce a scaling constant $\omega \in \mathbb{R}_{>0}$ such that $\frac{1}{\omega} U$ preserves the pixel values in this specific case.

We define the downsampling operation $\downsample\lft(\x_{\globlevel}\rgt)$ as 
\begin{equation}
    \downsample\lft(\x_{\globlevel}\rgt) = \frac{1}{\omega} \Pi_\lo \mathcal{W} \x_{\globlevel} =  \frac{1}{\omega} \hat{\x}_{\globlevel,\lo}.
\end{equation}

Introduced scaling keeps the range of pixel values after this operation similar to natural signals.
As an example, for Haar wavelet downsampling $\downsample\lft(\cdot\rgt)$ is just equal to average pooling.

From the calculations above, 
\begin{align}
p\lft(\downsample\lft(\x_{\globlevel}\rgt) \mid \x_0\rgt) 
&= \mathcal{N}\left(\left(1 - \globlevel\right) \cdot \frac{1}{\omega} U \x_0, \, \frac{\globlevel^2}{\omega^2} I\right) \nonumber \\
&= \mathcal{N}\left(\left(1 - \globlevel\right) \cdot \downsample\lft(\x_0\rgt), \, \frac{\globlevel^2}{\omega^2} I\right),
\label{eq:distribution_downsampled}
\end{align}
and, consequently, this distribution can be reparametrized as 
\begin{equation}
\downsample\lft(\x_{\globlevel}\rgt)  = \left(1 - \globlevel\right) \cdot \downsample\lft(\x_0\rgt) + \dfrac{\globlevel}{\omega} \varepsilon
\label{eq:distribution_downsampled_reparam}
\end{equation}
for $\varepsilon$ sampled from $\mathcal{N}\left(0, I\right).$

For the noisy signal $\x_{\globlevel}$ at the original resolution, we can calculate the signal-to-noise ratio as 
\begin{equation}
    \textrm{SNR}\lft[\x_{\globlevel} \mid \x_0\rgt] = \frac{\left\lVert \left(1-\globlevel\right) \x_0 \right\rVert^2}{\mathbb{E} \left\lVert \globlevel \epsilon \right\rVert^2} = \left(\frac{1-\globlevel}{\globlevel}\right)^2 \frac{\left\lVert \x_0 \right\rVert^2}{\mathbb{E} \left\lVert \epsilon \right\rVert^2}.
\end{equation}

For the downsampled noisy signal, from~\cref{eq:distribution_downsampled_reparam}
\begin{align}
    \textrm{SNR}\lft[\downsample\lft(\x_{\globlevel}\rgt) \mid \x_0\rgt] 
    &= \frac{\left\lVert \left(1-\globlevel\right) \downsample\lft(\x_0\rgt) \right\rVert^2}{\mathbb{E} \left\lVert \frac{\globlevel}{\omega} \varepsilon \right\rVert^2} \nonumber \\
    &= \omega^2 \left(\frac{1-\globlevel}{\globlevel}\right)^2  \frac{\left\lVert \downsample\lft(\x_0\rgt) \right\rVert^2}{\mathbb{E} \left\lVert \varepsilon \right\rVert^2} \nonumber \\
    &\approx \omega^2 \cdot \textrm{SNR}\lft[\x_{\globlevel} \mid \x_0\rgt].
\end{align}
Now, if we were to start a new forward diffusion process at the lower resolution, we would parametrize it as $\left(1 - \tau\right) \downsample\lft(\x_0\rgt) + \tau \eta $ for Gaussian random noise $\eta$ and noise level $\tau$.
To match signal to noise ratio of this process with \cref{eq:distribution_downsampled_reparam}, we need to solve the equation
\begin{equation}
    \left(\frac{1-\tau}{\tau}\right)^2 = \omega^2 \left(\frac{1-\globlevel}{\globlevel}\right)^2 
\end{equation}
which results in $\globlevel = \frac{\omega\tau}{1 + \left(\omega-1\right)\tau}.$
This coincides with the results obtained in the prior works~\cite{pmlr-v235-esser24a,hoogeboom_simple_2023,chen_importance_2023} in context of high-resolution diffusion models in pixel space.

If we denote by $i$ the number of downsampling operations applied one after another, we can rewrite the above equation as the relation between \emph{global noise levels} at scales $i$ and $i+1$, namely,
\begin{equation}
    \globlevel[i] = \frac{\omega\globlevel[i+1]}{1 + \left(\omega-1\right)\globlevel[i+1]}.
    \label{eq:glob_level_down}
\end{equation}
By `global' we mean that this noise levels are used in the forward diffusion equation.
We also will refer to the global noise level at highest resolution $\globlevel[0]$ as the \emph{natural noise level} $\snrlevel$.
Relation between global level $\globlevel[i]$ and natural noise level $\snrlevel$ is obtained with recursive application of the above equation.

\subsection{Upsampling}
\label{sec:upsampling}
Since we used the wavelet analysis matrix to define downsampling, we can now use the inverse operation to upsample the noisy signal $\x_{\globlevel}  \sim p\lft(\x_{\globlevel} \mid \x_0\rgt) = \mathcal{N}\lft(\left(1 - \globlevel\rgt) \x_0, \globlevel^2 I \right).$
For that purpose, we treat $\x_{\globlevel}$ as a low-frequency band and sample high-frequency bands independently from zero-centered Gaussian distribution $\mathcal{N}\lft(0, \nu^2 I\rgt)$.
The concatenated vector with all bands is then distributed as $\mathcal{N}\lft( \left(\begin{smallmatrix}
    \left(1 - \globlevel\right) \x_0 \\
    0
\end{smallmatrix}\right), \,
\left(\begin{smallmatrix}
    \globlevel^2 I & 0 \\
    0 & \nu^2 I
\end{smallmatrix}\right)
\rgt)$.
After synthesis matrix $\mathcal{W}^T$ is applied, the resulting vector $\x_{\globlevel,\nu}^{\uparrow}$ is distributed as
\begin{equation*}
    p\lft(\x_{\globlevel,\nu}^{\uparrow} \mid \x_0\rgt) = \mathcal{N}\left( \mathcal{W}^T \left(\begin{smallmatrix}
    \left(1 - \globlevel\right) \x_0 \\
    0
\end{smallmatrix}\right), \,
\mathcal{W}^T \left(\begin{smallmatrix}
    \globlevel^2 I & 0 \\
    0 & \nu^2 I
\end{smallmatrix}\right) \mathcal{W}
\right).
\end{equation*}
Since $\mathcal{W} = \left(U^T, \; V^T\right)^T$, we can rewrite the mean and covariance as 
\begin{align}
    \mathbb{E}\lft[\x_{\globlevel,\nu}^{\uparrow} \mid \x_0  \rgt]
    &= \left(1 - \globlevel\right) \mathcal{W}^T \left(\begin{smallmatrix}
     \x_0 \\
    0
    \end{smallmatrix}\right) \nonumber \\
    &=\left(1 - \globlevel\right)  \left(U^T, \, V^T\right) \left(\begin{smallmatrix}
    \x_0 \\
    0
    \end{smallmatrix}\right) \nonumber \\
    &= \left(1 - \globlevel\right) U^T \x_0,
\end{align}
\begin{align}
    \textrm{Cov}\lft[ \x_{\globlevel,\nu}^{\uparrow} \mid \x_0  \rgt]
    &= \mathcal{W}^T \left(\nu^2 I + 
    \left(\begin{smallmatrix}
    \left(\globlevel^2 - \nu^2\right) I & 0 \\
    0 & 0
    \end{smallmatrix}\right) 
    \right)\mathcal{W} \nonumber \\
    &= \nu^2 I + \left(U^T, \, V^T\right) 
    \left(\begin{smallmatrix}
    \left(\globlevel^2 - \nu^2\right) I & 0 \\
    0 & 0
    \end{smallmatrix}\right) \left(\begin{smallmatrix}
    U \\
    V
    \end{smallmatrix}\right) \nonumber \\
    &= \nu^2 I + \left(\globlevel^2 - \nu^2\right) U^T U.
\end{align}
To keep the magnitude of the mean similar to natural signals, we need to scale it by $\omega$ -- the opposite of what was required for downsampling.
We would like to match the obtained distribution with that of the forward diffusion process starting with $\omega U^T \x_0$ and parametrized as $\left(1-\tau\right) \cdot \omega U^T \x_0 + \tau \varepsilon$.
To achieve this, we additionally multiply the upscaled signal by a non-negative constant $r$ and solve the system of equations \wrt $r$, $\nu$, and $\tau$,
\begin{align}
    1 - \tau &= \left(1-\globlevel\right) r, \\
    \tau^2 I &= r^2 \omega^2 \left( \nu^2 I + \left(\globlevel^2 - \nu^2\right) U^T U \right).
\end{align}
The last equation can be rewritten as 
\begin{equation}
    \left(\tau^2 - r^2 \omega^2 \nu^2 \right) I = r^2 \omega^2 \left(\globlevel^2 - \nu^2\right) U^T U.
\end{equation}
Note that $U$ is a `wide' rectangular matrix, and therefore the rank of RHS is not greater than rank of LHS.
Thus, the equality can be fulfilled only if both sides are equal to 0, which leads to $\tau = r \omega \nu$ and $\nu = \globlevel$.
Consequently, 
\begin{align}
    r &= \frac{1}{1 + \left(\omega - 1\right)\globlevel}, \\
    \tau &= \frac{\omega \sigma}{1 + \left(\omega - 1\right)\globlevel}.
    \label{eq:glob_level_up}
\end{align}
We define two new functions of  a noisy signal $\x_{\globlevel}$, upsampling $\upsample$ and upsampling-and-renoising $\upsamplegauss$ as 
\begin{align}
    \upsample\lft(\x_{\globlevel}\rgt) &= \omega \x_{\globlevel,0}^{\uparrow}, \\ 
    \upsamplegauss\lft(\x_{\globlevel}\rgt) &= r \omega \x_{\globlevel,\globlevel}^{\uparrow} = \frac{\omega}{1 + \left(\omega - 1\right)\globlevel} \x_{\globlevel,\globlevel}^{\uparrow}. \label{eq:def_upsamplegauss}
\end{align}
From the derivations above, 
\begin{equation*}
    \upsamplegauss\lft(\x_{\globlevel}\rgt) = \left(1 - \tau\right) \upsamplegauss\lft(\x_{0}\rgt) + \tau \varepsilon.
\end{equation*}
With the scale index $i$ and global noise level notation introduced above, we can reformulate \cref{eq:glob_level_up} as
\begin{equation}
    \globlevel[i] = \frac{\omega \globlevel[i+1]}{1 + \left(\omega - 1\right)\globlevel[i+1]}.
\label{eq:glob_level_up}
\end{equation}
Note that this coincides with \cref{eq:glob_level_down}.
This leads to an important observation: both downsampling operation $\downsample\lft(\x_{\globlevel[i]}\rgt)$ and upsampling $\upsamplegauss\lft(\x_{\globlevel[i]}\rgt)$  applied to the noisy sample at scale $i$ do not change the natural noise level $\snrlevel$ associated with the global noise level $\globlevel[i]$.

\subsection{Downsampling + upsampling}
Consider the forward diffusion process at scale $i+1$ that goes from $\downsample\lft(\x_0\rgt)$ to the global noise level $\globlevel[i+1]$.
Then, after $\upsamplegauss$ being applied to this noisy sample, the result has the same marginal distribution as another diffusion process at scale $i$ with noise level $\globlevel[i]$ that started with $\upsamplegauss\circ\downsample\lft(\x_0\rgt)$.
But for the clean signal $\x_0$ at scale $i$ composition of introduced operations $\upsamplegauss \circ \downsample$ is equivalent to wavelet-based low-pass filtering.

This means that if the noise level $\globlevel[i]$ is high enough to turn the high-frequency part of $\x_{\globlevel[i]}$ into \iid noise, then with this (or larger) amount of noise the discrepancy between processes starting with $\x_0$ and $\upsamplegauss \circ \downsample\lft(\x_0\rgt)$ is negligible. 
Therefore, part of the backward diffusion process can run at lower resolution until the level $\globlevel[i+1]$, followed by upsampling $\upsamplegauss$ and backward process starting from $\globlevel[i]$.

In practice, noise levels suitable for switching between resolutions are calculated based on the average spectrum of the dataset, and therefore it is not guaranteed that for \emph{any} $\x_0$ the noise level $\globlevel[i]$ is high enough to destroy its high-frequency content.
Thus, it is more convenient to train a denoising model on inputs sampled from the downsampling-upsampling process rather than the original one.

%% file: supplementary/experiments.tex
\section{Additional experiments}
\label{sec:add_experiments}

\begin{table*}[t]
\centering
\caption{\textbf{Jenga results.} 
We measure the total latency of Video DiT calls for Jenga~\cite{zhang2025training} applied to Wan-DMD checkpoint with 2 sampling steps.
For comparison, our Wan-PPF-DMD model with 2-2-1 schedule achieves a total transformer latency of 810 ms (standard deviation < 10 ms) at the same output video resolution.
Beyond offering a better quality–efficiency trade-off, pyramidal models benefit from a static computational graph, which greatly simplifies deployment.
}
\label{tab:jenga}
\resizebox{0.75\textwidth}{!}{%
\input{tables/jenga}
}
\vspace{-1em}
\end{table*}

Both PyramidalFlow and PPF can be viewed as special cases of token reduction strategies during inference. 
Token reduction techniques are well established in the community, with several recent methods proposed, such as ToMe~\cite{bolya2023tomesd}, ADAPTOR~\cite{Peruzzo_2025_CVPR}, RLT~\cite{choudhury2024dont}, and vid-TLDR~\cite{Choi_2024_CVPR}. 
However, these approaches typically rely on heuristics based on mutual token similarity. 
Consequently, their speed-up procedure is \emph{dynamic}, depending on factors such as prompt complexity and initial latent noise.
In contrast, pyramidal approaches can be represented as \emph{static} computational graphs (one per stage), which greatly simplifies deployment, particularly in resource-constrained environments~\cite{anonymous2025neodragon}.

Despite this fundamental difference, we include a comparison with the recent Jenga method~\cite{zhang2025training}. 
Jenga employs a dynamic sparse attention mask based on token similarity and leverages space-filling curves for GPU-friendly implementation. 
We apply Jenga-Base (without progressive resolution) and Jenga-Turbo (with half of the steps performed on downsampled latents) on top of our Wan-DMD checkpoint and measure performance and latency using 2 sampling steps. 
For Jenga-Turbo, we tried downsampling factors of 0.75 and 0.5.
Note that the original Jenga implementation does not apply downsampling along the temporal dimension, even when using progressive resolution.
We found that adding it to spatial downsampling indeed leads to noticeable drop of quality.
Our default compiler settings (see~\cref{sec:details}) did not work for Jenga.
For that reasons, we adjusted the settings to \texttt{fullgraph=False} and \texttt{mode='max-autotune-no-cudagraphs'}. 
Results  are reported in~\cref{tab:jenga}. 
For details on \emph{AttenCarve} hyperparameters, please refer to the work of \citet{zhang2025training}. 
TeaCache~\cite{Liu_2025_CVPR} was not used in these experiments.

While Jenga obtains good VBench and VBench-2.0 scores, we found that videos produced by these models often suffer from abrupt scene changes, incoherent motion, and `episodic' behavior.
These artifacts are more pronounced than in case of original Wan-DMD model or our pyramidal modifications.
Please, refer to the supplementary videos for comparison.
Overall, our experiments demonstrate that pyramidization results in better quality-efficiency trade-off.

%% file: tables/jenga.tex
\begin{tabular}{ccccccccc}
\toprule
\multirow{2}{*}{Model} & \multicolumn{2}{c}{AttenCarve}&  & \multicolumn{2}{c}{ProRes of 1st step}  & \multirow{2}{*}{\makecell{Latency, \\ms~$\downarrow$}} & \multicolumn{2}{c}{Total score~$\uparrow$}  \\%
\cmidrule{2-3}\cmidrule{5-6}\cmidrule{8-9} 
& $k$ & $p$ & & factor & temporal downsampling &   & VBench & VBench-2.0 \\
\midrule
\multirow{4}{*}{Jenga-Base} & 0.05 &  0.3 && $1\times$ & \nomark & \num{1211} $\pm$ 199 & 81.25 & 54.30 \\
& 0.1 &  0.3 && $1\times$ & \nomark & \num{1278} $\pm$ \num{205} & 83.52 & 52.84 \\
& 0.15 & 0.3 && $1\times$ & \nomark & \num{1297} $\pm$ \num{304} & 83.54 & \num{53.28} \\
& 0.15 & 0.9 && $1\times$ & \nomark & \num{1680} $\pm$ \num{201} & 83.13 & 57.19 \\
\midrule
\multirow{3}{*}{Jenga-Turbo} & 0.1 & 0.3 && $0.75\times$ & \nomark & \num{1089} $\pm$ \num{145} & 83.38 & 53.94 \\
 & 0.1 & 0.3 && $0.5\times$ & \nomark & \num{932} $\pm$ \num{114} & 82.98 & 53.46 \\
 & 0.1 & 0.3 && $0.5\times$ & \yesmark & \num{865} $\pm$ \num{350} & 81.23 & 52.30 \\
\bottomrule
\end{tabular}

%% file: supplementary/details.tex
\section{Additional details}
\label{sec:details}

In this section we provide training and evaluation details useful for reproducing our experiments.

\paragraph{Handling spatiotemporal resolution}
In the default setting, Wan model operates with 21 latent frame at stage $i=0$. 
This requires special treatment for upsampling and downsampling along the temporal axis.
For our final experiments on PyramidalWan, we opted for separate handling of the first frame.
Namely, during the forward diffusion process its global noise level $\globlevel_\textrm{first}$ is different from the global noise level of the rest of the frames $\globlevel_\textrm{rest}$, but both of them correspond to the same natural level $\snrlevel$.
This is due to the fact that relation between global and natural levels depends on the scaling factor $\omega$ (see~\cref{eq:glob_level_up}), which differs for 2D and 3D cases.
Similarly, during the upscaling operation $\upsamplegauss$ first frame is upsampled only spatially, while the rest of the frames --- spatiotemporally, both according to~\cref{eq:def_upsamplegauss}.
The derivative of the noised signal \wrt the global level is computed separately for the first and the other frames.
Since RoPE positional embeddings~\cite{heo_rotary_2025} used in Wan could potentially make it challenging for the network to handle this special nature of the first frame, we added a special learnable embedding vector to all the tokens of the first frame after the patchification layer.

For PPF, increased kernel size of the patchification layer also introduces inconsistency with the original shape of the latent video tensor.
Therefore, for stages $i=1$ and $i=2$ we apply trilinear interpolation to upsample the original tensor to the minimal shape that allows to apply the patchifier.
\Eg, for stage 2 we resized from $21 \times 60 \times 104$ to $24 \times 64 \times 104$, thus enabling patchification with kernel size $4 \times 8 \times 8$.
After the unpatchification layer, the last operation of VideoDiT, trilinear interpolation is applied again to resize the tensor to the original shape.
Note that for PPF we treat the first frame in the same way as other frames.

\paragraph{Wan-DMD}
For DMD distillation of the original model, fake score network was updated twice per each update of the student model.
The learning rate of AdamW optimizer for the student was \num{1e-5}, and for fake score \num{5e-6}.
Noise levels for the student's inputs were selected uniformly from the set $\left\{ 0.0050, 0.7149, 0.9092, 1\right\}$, corresponding to uniform selection of \emph{timesteps} between 1 and 1000 with the consequent application of \emph{shift}~\cite{pmlr-v235-esser24a} equal to 5.
For fake score network, noise levels were selected uniformly, with further application of shift of 5.
For teacher's outputs, classifier-free guidance was applied with scale value of 5. 
We used the negative prompt~\cite{NEURIPS2020_49856ed4} \emph{``Bright tones, overexposed, static, blurred details, subtitles, style, works, paintings, images, static, overall gray, worst quality, low quality, JPEG compression residue, ugly, incomplete, extra fingers, poorly drawn hands, poorly drawn faces, deformed, disfigured, misshapen limbs, fused fingers, still picture, messy background, three legs, many people in the background, walking backwards''}.

\paragraph{Wan-Adv}
The discriminator head comprises two branches: a \textit{spatial} branch and a \textit{temporal} branch. Both branches share a common processing pipeline: input reshaping, initial convolution, one or more ResNet blocks, SiLU activation, and a final convolution, followed by restoring the original video layout.
The spatial branch operates on frame-level structure by flattening the temporal dimension into the batch, $(b,t,c,h,w) \mapsto (b \cdot t,c,h,w)$, and applies 2D convolutions and 2D ResNet blocks to capture intra-frame details.
The temporal branch focuses on temporal dynamics by flattening spatial dimensions, $(b,t,c,h,w) \mapsto (b \cdot h \cdot w,c,t,1,1)$, and uses 3D convolutions with kernel size $3 \times 1 \times 1$ and temporal ResNet blocks to model inter-frame relationships.
Weights are initialized with Xavier normal. The final convolution is zero-initialized for more stable adversarial training.
For adversarial distillation, the discriminator head (with the backbone feature extractor kept frozen) is updated four times more frequently than the generator (student model). We use the AdamW optimizer with a learning rate of \num{1e-4} for the discriminator head and \num{1e-5} for the student model. Noise levels for student model distillation are sampled uniformly from the set $\{0.25, 0.5, 0.75, 1\}$.

\paragraph{PyramidalWan}
To split the natural noise scale between stages, we conducted the spectral analysis of latents produced by the encoder of WanVAE.
Our analysis follows the same procedure as that described, among others, by \citet{dieleman2024spectral,anonymous2025scalewise}.
Based on the results we set the `cleaner' natural leves as follows:
$\snrlevel^{\left(1\right)}_c = 0.5858$, $\snrlevel^{\left(2\right)}_c = 0.9412$.

For training of our pyramidal flow matching model, for each stage $i$ we sampled the natural noise level uniformly, 
\begin{align}
    u &\sim \textrm{Uni}\lft(0, 1\rgt), \label{eq:uniform_u} \\
    \snrlevel^{\upi} &= \snrlevel^{\upi}_c + u \cdot \left(\snrlevel^{\upi}_n - \snrlevel^{\upi}_c\right), \label{eq:snr_level_from_u} 
\end{align}
where natural levels $\snrlevel^{\upi}_c$ and $\snrlevel^{\upi}_n$ correspond to the global boundary levels $\globlevel[i]_c$ and $\globlevel[i]_n$ with 3D scaling factor $\omega$, see~\cref{eq:glob_level_up}.

For DMD-PT pipeline, student's $u$ (see~\cref{eq:snr_level_from_u}) was selected uniformly from the set $\left\{0.25, 0.5, 0.75, 1 \right\}$, while for the fake score it was sampled according to~\cref{eq:uniform_u}.
In DMD-OT training, we sampled $u$ \wrt \cref{eq:uniform_u} and applied shift 5 afterwards.
For the fake score (since it is initialized with the original Wan model), natural noise level $\snrlevel'$ was sampled from $\textrm{Uni}\lft(0, 1\rgt) $, and shift 5 was applied after that.

PyramidalWan has been trained with learning rate of \num{1e-5}.
For PyramidalWan-DMD, optimizers had the same hyperparameters as for Wan-DMD.
To mix videos of different spatiotemporal resolution in the same batch, we flattened all the tokens into a single 1D sequence and applied sparse self-attention and cross-attention masks using FlexAttention~\cite{dong2025flexattention}.

For adversarial distillation of PyramidalWan (Adv-OD and Adv-PD), we follow the same hyperparameters as Wan-Adv. The only difference is that the noise levels for the student model are adjusted at each stage according to~\cref{eq:snr_level_from_u}.

\paragraph{Wan-PPF}
For PPF models, we used the same hyperparameters as for the training of the full Wan models.
Noise levels were split between stages in the same way as for PyramidalWan.

\paragraph{Sampling from trained models}
For all the models, both multi-step and few-step, we used \texttt{FlowMatchEulerDiscreteScheduler} from the diffusers library for sampling~\cite[v0.33.0]{von-platen-etal-2022-diffusers}.
For Wan-DMD, noise levels for 4-step sampling were taken from the set $\left\{1, 0.9097, 0.7173, 0.0244 \right\}$, and for 2-step sampling from $\left\{1, 0.8347 \right\}$.
For PyramidalWan-DMD with 2-2-1 schedule, natural noise levels were selected as follows: $\snrlevel^{\left(2\right)} \in \left\{1, 0.9863 \right\},$ $\snrlevel^{\left(1\right)} \in \left\{0.9412, 0.8645 \right\},$ $\snrlevel^{\left(0\right)} \in \left\{0.5858 \right\}.$ 
Same schedule was used for Wan-PPF-DMD.
For the multi-step flow matching models, we used classifier-free guidance scale of 5.

\paragraph{Measurements}
To estimate the computational cost of various models, we calculated FLOPs using DeepSpeed library~\cite[v0.14.2]{deepspeed_2020}.
For latency measurements, models were compiled separately for each resolution with the PyTorch compiler~\cite[v2.7.0]{pytorch_neurips_2019} on H100 GPU.
The configuration of the compiler was set as follows: \texttt{dynamic=False}, \texttt{fullgraph=True}, \texttt{mode="max-autotune"}.
For VBench and VBench-2.0 evaluation of all models mentioned in the paper, generated videos were saved using TorchVision's~\cite[v0.22.0]{torchvision2016} \texttt{torchvision.io.write\_video} function with default parameters.
We used \emph{GPT}\footnote{\url{https://github.com/Vchitect/VBench/blob/7bcb8691c49426ac30544456d19a234d971722e6/prompts/augmented_prompts/gpt_enhanced_prompts/all_dimension_longer.txt}} set of extended prompts to evaluate VBench and \emph{Wanx}\footnote{\url{https://github.com/Vchitect/VBench/blob/8270c9e54eb56de9a589ec351f4ff3c4e0ab3dfd/VBench-2.0/prompts/prompt_aug/Wanx_full_text_aug.txt}} set for VBench-2.0.

\input{tables/vbench_full_agg}
\input{tables/vbench2_full_agg}

%% file: tables/vbench_full_agg.tex
\begin{table*}[t]
\centering
\caption{\textbf{VBench scores.} 
}
\label{tab:vbench_full_aggregated}
\resizebox{\textwidth}{!}{%
\begin{tabular}{lcccccccccc}
\toprule
\textbf{\makecell[c]{Models}} & \textbf{\makecell[c]{Subject\\Consistency}} & \textbf{\makecell[c]{Background\\Consistency}} & \textbf{\makecell[c]{Temporal\\Flickering}} & \textbf{\makecell[c]{Motion\\Smoothness}} & \textbf{\makecell[c]{Dynamic\\Degree}} & \textbf{\makecell[c]{Aesthetic\\Quality}} & \textbf{\makecell[c]{Imaging\\Quality}} & \textbf{\makecell[c]{Object\\Class}} & \textbf{\makecell[c]{Multiple\\Objects}} & \textbf{\makecell[c]{Human\\Action}} \\
\midrule
Wan2.1-1.3B (50 steps) & 93.07 & 95.21 & 99.35 & 98.03 & 69.17 & 65.20 & 65.07 & 88.78 & 72.07 & 96.40 \\
Wan2.1-1.3B (25 steps) & 92.06 & 95.43 & 99.30 & 97.13 & 69.44 & 62.91 & 62.28 & 85.89 & 66.40 & 96.99 \\
PyramidalWan (20-20-10) & 97.02 & 97.38 & 99.44 & 98.68 & 44.44 & 66.13 & 65.69 & 95.27 & 84.47 & 95.60 \\
\midrule
Wan-Adv (4 steps) & 95.62 & 95.44 & 98.47 & 98.87 & 71.39 & 63.37 & 65.79 & 91.33 & 71.91 & 95.00 \\
Wan-Adv (2 steps)  & 95.97 & 95.31 & 98.58 & 98.93 & 64.17 & 63.68 & 66.24 & 90.35 & 69.91 & 93.80 \\
Wan-Adv (1 step)  & 95.04 & 94.71 & 98.59 & 98.80 & 39.44 & 63.23 & 66.08 & 88.26 & 68.61 & 92.60 \\
Wan-DMD (4 steps)  & 94.57 & 94.34 & 97.91 & 97.78 & 85.83 & 66.66 & 67.43 & 92.94 & 77.76 & 96.20 \\
Wan-DMD (2 steps)  & 95.81 & 94.92 & 98.27 & 98.28 & 64.44 & 66.87 & 68.42 & 95.11 & 84.45 & 97.40 \\
Wan-DMD (1 step)  & 92.95 & 92.94 & 98.55 & 98.44 & 46.94 & 59.96 & 66.89 & 84.43 & 64.65 & 95.60 \\
\midrule
PyramidalWan-Adv-OD (2-2-1) & 97.63 & 97.17 & 98.96 & 98.49 & 45.83 & 66.48 & 69.94 & 94.21 & 79.80 & 94.40 \\
PyramidalWan-Adv-PD (2-2-1) & 96.39 & 96.71 & 99.23 & 98.79 & 56.94 & 65.08 & 67.82 & 93.94 & 72.84 & 91.80 \\
Wan-PPF-DMD (2-2-1) & 96.18 & 94.34 & 98.37 & 98.80 & 49.72 & 65.30 & 69.35 & 94.95 & 85.40 & 94.60 \\
PyramidalWan-DMD-OT (2-2-1) & 97.56 & 96.74 & 97.66 & 98.04 & 43.89 & 69.96 & 71.14 & 95.97 & 84.24 & 95.00 \\
PyramidalWan-DMD-PT$^{*}$ (2-2-1) & 98.06 & 96.98 & 99.34 & 99.07 & 31.39 & 68.07 & 69.24 & 95.44 & 85.18 & 94.20 \\
\bottomrule
\end{tabular}
}

\vspace{1em} 

\resizebox{\textwidth}{!}{%
\begin{tabular}{lccccccccc}
\toprule
\textbf{\makecell[c]{Models}} & \textbf{\makecell[c]{Color}} & \textbf{\makecell[c]{Spatial\\Relationship}} & \textbf{\makecell[c]{Scene}} & \textbf{\makecell[c]{Appearance\\Style}} & \textbf{\makecell[c]{Temporal\\Style}} & \textbf{\makecell[c]{Overall\\Consistency}} & \textbf{\makecell[c]{Quality\\Score}} & \textbf{\makecell[c]{Semantic\\Score}} & \textbf{\makecell[c]{Total\\Score}} \\
\midrule
Wan2.1-1.3B (50 steps) & 83.20 & 75.46 & 54.56 & 22.82 & 25.78 & 26.99 & 83.47 & 78.57 & 82.49 \\
Wan2.1-1.3B (25 steps)  & 83.11 & 67.94 & 49.64 & 23.69 & 24.90 & 26.28 & 82.09 & 76.02 & 80.87 \\
PyramidalWan (20-20-10)  & 88.51 & 77.37 & 55.12 & 21.96 & 24.92 & 26.48 & 83.36 & 80.70 & 82.83 \\
\midrule
Wan-Adv (4 steps) & 84.70 & 75.78 & 51.21 & 22.01 & 24.22 & 26.19 & 84.06 & 77.39 & 82.72 \\
Wan-Adv (2 steps) & 86.22 & 74.16 & 51.80 & 21.71 & 24.15 & 26.08 & 83.74 & 76.82 & 82.35 \\
Wan-Adv (1 step) & 84.74 & 75.98 & 50.51 & 21.30 & 23.85 & 25.86 & 81.38 & 75.85 & 80.28 \\
Wan-DMD (4 steps) & 80.93 & 71.82 & 51.90 & 21.60 & 25.24 & 26.59 & 84.71 & 77.86 & 83.34 \\
Wan-DMD (2 steps) & 86.04 & 78.51 & 51.95 & 21.65 & 25.32 & 26.82 & 84.00 & 80.41 & 83.28 \\
Wan-DMD (1 step) & 84.99 & 68.52 & 47.51 & 22.15 & 24.60 & 26.07 & 80.63 & 74.75 & 79.45 \\
\midrule
PyramidalWan-Adv-OD (2-2-1) & 85.82 & 80.77 & 51.73 & 20.96 & 24.34 & 25.66 & 83.94 & 78.74 & 82.90 \\
PyramidalWan-Adv-PD (2-2-1) & 80.99 & 70.83 & 54.17 & 20.10 & 24.19 & 26.05 & 84.20 & 76.07 & 82.57 \\
Wan-PPF-DMD (2-2-1) & 88.90 & 79.48 & 51.85 & 20.46 & 24.88 & 26.17 & 83.04 & 79.80 & 82.39 \\
PyramidalWan-DMD-OT (2-2-1) & 83.96 & 82.18 & 50.77 & 22.22 & 24.09 & 25.90 & 83.63 & 79.80 & 82.86 \\
PyramidalWan-DMD-PT$^{*}$ (2-2-1) & 81.26 & 82.14 & 52.54 & 21.35 & 24.93 & 26.35 & 83.46 & 79.75 & 82.72 \\
\bottomrule
\end{tabular}
}
\end{table*}

%% file: tables/vbench2_full_agg.tex
\begin{table*}[t]
\centering
\caption{\textbf{VBench-2.0 scores.} 
}
\label{tab:vbench2_full_aggregated}
\resizebox{\textwidth}{!}{%
\begin{tabular}{lcccccccccccc}
\toprule
\textbf{\makecell[c]{Models}} & \textbf{\makecell[c]{Camera\\Motion}} & \textbf{\makecell[c]{Complex\\Landscape}} & \textbf{\makecell[c]{Complex\\Plot}} & \textbf{\makecell[c]{Composition}} & \textbf{\makecell[c]{Diversity}} & \textbf{\makecell[c]{Dynamic\\Attribute}} & \textbf{\makecell[c]{Dynamic\\Spatial\\Relationship}} & \textbf{\makecell[c]{Human\\Anatomy}} & \textbf{\makecell[c]{Human\\Clothes}} & \textbf{\makecell[c]{Human\\Identity}}  & \textbf{\makecell[c]{Human\\Interaction}}  & \textbf{\makecell[c]{Instance\\Preservation}} \\
\midrule
Wan2.1-1.3B (50 steps) & 49.08 & 72.44 & 31.99 & 61.70 & 48.42 & 40.80 & 97.99 & 11.31 & 25.12 & 85.96 & 63.50 & 16.44 \\
Wan2.1-1.3B (25 steps) & 96.10 & 47.62 & 10.91 & 73.13 & 62.90 & 25.12 & 19.11 & 83.04 & 4.02 & 79.33 & 48.65 & 71.33 \\
PyramidalWan (20-20-10) & 99.06 & 18.68 & 12.85 & 57.26 & 74.39 & 31.88 & 17.78 & 95.91 & 37.40 & 82.71 & 44.79 & 72.00 \\
\midrule
Wan-Adv (4 steps) & 49.07 & 14.44 & 9.54 & 50.34 & 51.59 & 38.46 & 27.54 & 88.14 & 96.73 & 71.88 & 67.67 & 83.63 \\
Wan-Adv (2 steps) & 44.14 & 17.11 & 9.28 & 48.75 & 45.39 & 39.56 & 26.57 & 87.12 & 95.85 & 65.18 & 60.67 & 82.46 \\
Wan-Adv (1 step) & 40.43 & 17.56 & 9.65 & 45.51 & 31.51 & 41.39 & 27.05 & 86.07 & 85.52 & 67.20 & 59.00 & 78.36 \\
Wan-DMD (4 steps) & 99.48 & 51.28 & 10.92 & 76.34 & 60.84 & 35.27 & 16.67 & 79.53 & 24.51 & 88.08 & 48.53 & 71.67 \\
Wan-DMD (2 steps) & 37.96 & 20.22 & 13.58 & 48.70 & 45.30 & 43.96 & 31.40 & 87.25 & 100.00 & 54.94 & 75.00 & 85.96 \\
Wan-DMD (1 step) & 89.81 & 54.21 & 10.23 & 74.22 & 73.15 & 30.92 & 16.89 & 80.12 & 2.21 & 69.53 & 47.13 & 62.00 \\
\midrule
PyramidalWan-Adv-OD (2-2-1) & 23.77 & 16.22 & 7.87 & 39.79 & 49.82 & 13.92 & 24.15 & 61.12 & 100.00 & 84.09 & 62.00 & 93.57 \\
PyramidalWan-Adv-PD (2-2-1) & 28.40 & 15.56 & 9.20 & 42.99 & 47.02 & 21.61 & 28.99 & 88.56 & 98.60 & 75.78 & 59.33 & 89.47 \\
Wan-PPF-DMD (2-2-1) & 31.79 & 16.67 & 10.36 & 48.62 & 31.46 & 12.45 & 24.64 & 88.23 & 100.00 & 86.61 & 66.67 & 89.47 \\
PyramidalWan-DMD-OT (2-2-1) & 24.69 & 15.56 & 9.00 & 45.56 & 54.52 & 25.64 & 30.92 & 45.52 & 100.00 & 86.98 & 73.67 & 91.23 \\
PyramidalWan-DMD-PT$^{*}$ (2-2-1) & 22.53 & 13.56 & 10.39 & 42.81 & 26.80 & 21.98 & 29.95 & 63.55 & 100.00 & 80.60 & 75.00 & 95.32 \\\bottomrule
\end{tabular}
}

\vspace{1em} 

\resizebox{\textwidth}{!}{%
\begin{tabular}{lcccccccccccc}
\toprule
\textbf{\makecell[c]{Models}} & \textbf{\makecell[c]{Material}} & \textbf{\makecell[c]{Mechanics}} & \textbf{\makecell[c]{Motion\\Order\\Understanding}} & \textbf{\makecell[c]{Motion\\Rationality}} & \textbf{\makecell[c]{Multi-View\\Consistency}} & \textbf{\makecell[c]{Thermotics}} & \textbf{\makecell[c]{Creativity\\Score}} & \textbf{\makecell[c]{Commonsense\\Score}} & \textbf{\makecell[c]{Controllability\\Score}} & \textbf{\makecell[c]{Human\\Fidelity\\Score}} & \textbf{\makecell[c]{Physics\\Score}} & \textbf{\makecell[c]{Total\\Score}} \\
\midrule
Wan2.1-1.3B (50 steps) & 9.62 & 80.63 & 71.67 & 69.44 & 49.05 & 32.10 & 48.73 & 63.38 & 33.96 & 80.71 & 53.30 & 56.02 \\
Wan2.1-1.3B (25 steps) & 41.95 & 67.57 & 50.34 & 36.70 & 34.26 & 64.08 & 49.49 & 62.50 & 35.01 & 79.44 & 52.20 & 55.73 \\
PyramidalWan (20-20-10) & 32.76 & 59.22 & 44.50 & 23.91 & 21.60 & 53.68 & 44.64 & 64.33 & 28.39 & 85.38 & 51.89 & 54.93 \\
\midrule
Wan-Adv (4 steps) & 64.66 & 75.74 & 21.21 & 31.03 & 0.00 & 61.87 & 50.96 & 57.33 & 32.56 & 85.58 & 50.57 & 55.40 \\
Wan-Adv (2 steps) & 62.07 & 73.08 & 23.23 & 34.48 & 18.90 & 63.38 & 47.07 & 58.47 & 31.51 & 82.72 & 54.36 & 54.82 \\
Wan-Adv (1 step) & 63.16 & 73.88 & 14.81 & 30.46 & 0.00 & 60.14 & 38.51 & 54.41 & 29.99 & 79.60 & 49.29 & 50.36 \\
Wan-DMD (4 steps) & 36.78 & 71.30 & 53.20 & 32.32 & 36.42 & 64.75 & 50.87 & 58.16 & 36.36 & 82.80 & 59.23 & 57.48 \\
Wan-DMD (2 steps) & 68.50 & 72.99 & 37.37 & 39.66 & 12.58 & 68.79 & 47.00 & 62.81 & 37.07 & 80.73 & 55.72 & 56.67 \\
Wan-DMD (1 step) & 38.51 & 72.95 & 30.21 & 30.98 & 42.28 & 70.63 & 38.67 & 59.31 & 35.36 & 77.49 & 55.00 & 53.17 \\
\midrule
PyramidalWan-Adv-OD (2-2-1) & 60.61 & 58.33 & 12.12 & 28.16 & 36.12 & 49.64 & 44.80 & 60.86 & 22.86 & 81.74 & 51.17 & 52.29 \\
PyramidalWan-Adv-PD (2-2-1) & 66.67 & 69.53 & 17.85 & 34.48 & 9.79 & 58.04 & 45.00 & 61.98 & 25.85 & 87.65 & 51.01 & 54.30 \\
Wan-PPF-DMD (2-2-1) & 59.62 & 61.90 & 10.77 & 36.21 & 12.78 & 57.64 & 40.04 & 62.84 & 24.76 & 91.61 & 47.98 & 53.45 \\
PyramidalWan-DMD-OT (2-2-1) & 64.49 & 62.32 & 23.91 & 35.63 & 39.33 & 60.99 & 50.04 & 63.43 & 29.05 & 77.50 & 56.78 & 55.36 \\
PyramidalWan-DMD-PT$^{*}$ (2-2-1) & 62.07 & 62.99 & 25.93 & 31.03 & 17.67 & 60.96 & 34.81 & 63.18 & 28.48 & 81.38 & 50.92 & 51.75 \\
\bottomrule
\end{tabular}
}
\end{table*}